\documentclass{article}
\usepackage{pp2020}
\usepackage{hyperref}
\usepackage[T1]{fontenc}
\usepackage[utf8]{inputenc}

\usepackage{textcomp,marvosym}

\begin{document}

\title{Semantics of European poetry is shaped by conservative forces: The relationship between poetic meter and meaning in accentual-syllabic verse}
\authorrunning{Semantics of European poetry is shaped by conservative forces}
\titlerunning{Semantics of European poetry is shaped by conservative forces}
\author{
    \ppauthname{Artjoms Šeļa}
    \ppaffil{Institute of Polish Language, Polish Academy of Sciences, Krakow, Poland}
    \ppaffil{University of Tartu, Tartu, Estonia}
    \ppemail{artjoms.sela@ijp.pan.pl}
    \pporcid{0000-0002-2272-2077}
    \and
    \ppauthname{Petr Plecháč}
    \ppaffil{Institute of Czech Literature, Czech Academy of Sciences, Prague, Czechia}
    \ppemail{plechac@ucl.cas.cz}
    \pporcid{0000-0002-1003-4541}
    \and
    \ppauthname{Alie Lassche}
    \ppaffil{Leiden University, Leiden, Netherlands}
    \ppemail{a.w.lassche@hum.leidenuniv.nl}  
    \pporcid{0000-0002-7607-0174}
}
\date{}
\maketitle 


\begin{abstract}

\noindent Recent advances in cultural analytics and large-scale computational studies of art, literature and film often show that long-term change in the features of artistic works happens gradually. These findings suggest that conservative forces that shape creative domains might be underestimated. To this end, we provide the first large-scale formal evidence of the persistent association between poetic meter and semantics in 18-19th European literatures, using Czech, German and Russian collections with additional data from English poetry and early modern Dutch songs. Our study traces this association through a series of clustering experiments using the abstracted semantic features of 150,000 poems. With the aid of topic modeling we infer semantic features for individual poems. Texts were also lexically simplified across collections to increase generalizability and decrease the sparseness of word frequency distributions. Topics alone enable recognition of the meters in each observed language, as may be seen from highly robust clustering of same-meter samples (median Adjusted Rand Index between 0.48 and 1 across traditions). In addition, this study shows that the strength of the association between form and meaning tends to decrease over time. This may reflect a shift in aesthetic conventions between the 18th and 19th centuries as individual innovation was increasingly favored in literature. Despite this decline, it remains possible to recognize semantics of the meters from past or future, which suggests the continuity of semantic traditions while also revealing the historical variability of conditions across languages. This paper argues that distinct metrical forms, which are often copied in a language over centuries, also maintain long-term semantic inertia in poetry. Our findings, thus, highlight the role of the formal features of cultural items in influencing the pace and shape of cultural evolution.

\end{abstract}

\section*{Introduction}

Recent advances in cultural analytics \cite{manovich_cultural_2020} and large-scale computational studies of creative domains such as art, literature and film provide an increasing evidence that change in features of artistic works happen gradually over extended periods of time. This can be seen in lexical choices in fiction \cite{heuser_quantitative_2012, morin_birth_2017}, writing styles \cite{hughes_quantitative_2012,storey_like_2020}, the shortening of cinematic shot lengths \cite{cutting_quicker_2011}, the long-term recognizability of literary genres \cite{underwood_distant_2019} and literary aesthetic choices \cite{underwood_2015}. This picture is puzzling because it suggests continuity and slow global processes in traditions that scholars have often perceived as volatile fields of innovation, competition and constant conflict of elites \cite{tynianov_permanent_2019,simmel_fashion_1904}. The contrary evidence for a punctuated equilibrium pattern of cultural evolution \cite{mauch_evolution_2015} or a cyclical turn-around of styles \cite{yarkho_speech_2019,peterson_cycles_1975,klimek_fashion_2019}  are also abundant, but a question remains: Are we underestimating the continuity and conservative forces at work in cultural traditions associated with creative freedom? This study asks this question about a practice which tends to be imagined as extremely individualistic: the composition of poetry. We apply a data-driven semantic analysis to the formal characteristics of texts across several languages. Our goal is to address one of the fundamental issues in versification studies: the connection between form and meaning.

Modern poetry is often seen as a space of boundless innovation and individualized self-expression. However, there is at least one aspect of poetry that is defined by conservative persistence over centuries and millennia: poetic meter. Poetic meter is rarely invented individually; rather it is a pre-existing prosodic pattern that usually arrives in the hands of a poet after long unbroken chains of usage in local and global traditions. These forms are not simply differently shaped pieces of the same blank page on which a poem is written: they are affected by what was previously written in these meters and by additional signals that poetic forms accumulate and carry over time. This persistence of formal features in poetry invites us to shift our focus from individuals to meters as cultural items that participate in long transmission chains \cite{morin_how_2015} that diffuse them far and wide.

Research into historical poetics and metrics strongly suggests that meter is not agnostic to meaning \cite{gasparov_metr_1999, tarlinskaja_meter_1986, tarlinskaja_meter_1989, cervenka_z_1991}. In oral traditions, a difference in meter was often functional: it supported genre diversification between the poles of epic and lyric poetry \cite{gasparov_history_1996}. The rise of written and printed media allowed more  diverse poetic metrical forms and expanded their sources. New forms could emerge from the standardization and restructuring of cultural borrowings: foreign traditions, classical Latin and Greek examples, adaptations of local oral versification systems. Through their usage, mnemonic capacities and generic conventions, metrical forms allegedly maintained fuzzy but distinct semantic traditions that were reproduced and updated by generations of poets. This theory of a relationship between a meter and its meaning is known as the ``semantic halo of meter''.

In this study, we aim to computationally test the presence of the semantic halo across several modern European poetic traditions  (18-20th c.). Most of the evidence for a semantic halo comes from sporadic informal studies of a few European (mostly Slavic) literatures \cite{gasparov_history_1996,tarlinskaja_meter_1986,pszczolowska_slowianska_1988, cervenka_z_1991,dobrzynska_verse_2014}. Several recent attempts to formalize this concept have succeeded in showing some lexical differences between metrical forms in individual traditions \cite{piperski_semantic_2017, orekhov_bashkirskii_2019} while also broadly confirming the presence of the semantic halo in 18th-to-20th century Russian verse \cite{sela_weak_2020}. This growing body of evidence suggests that the mechanism which binds form and meaning is widespread and could potentially be universal. However, a reliance on close reading, a lack of formalization and the sporadic nature of the research to date have limited scholars' ability to generalize about the semantic halo, its nature and historical dynamics.

The current study relies on several assumptions about the nature of poetic form and its relationship to verse semantics. First, we assume that poetry is a socially learned practice that is subject to cultural evolutionary forces \cite{boyd_culture_1988,mesoudi_cultural_2011,el_mouden_cultural_2014}. Poetic formal features such as meter and rhyme are reproduced and developed through a copying process that may be influenced by various factors, or biases. For example, acquisition and popularity of meters might depend on their intrinsic features, or fit to a language prosody (stress-timed languages often adopt accentual-syllabic meters). It also might be completely driven by cultural prestige, like medieval and early modern Latin hexameter: despite not perceiving any distinction between long and short vowels anymore, scholars and students continued to write ``correct'' quantitative hexameters for centuries, relying only on memorized rules \cite{gasparov_history_1996}.

Second, we assume that the effect of the semantic halo emerges through the copying and transmission of specific meters. We suppose that the semantics of a meter  depends entirely on the historical environment (i.e. form and meaning have an arbitrary and symbolic relationship) and not dependent on intrinsic features of form (iconic relationship \cite{dobrzynska_verse_2014}). This implies that meter has a transmission-limiting role: semantic features are likely to be carried over from an existing group of texts in a given form when that form is reproduced in a new poem. 

Given these two main assumptions (i.e. the social transmission and structural limitations), we argue that any poetic tradition that allows for structurally distinct poetic forms to exist over an extended time should exhibit the semantic halo effect. This study, however, seeks to find the effect mainly in three closely related European accentual-syllabic (AS) traditions: Czech, German and Russian. In addition, we look for further evidence in English poetry and early modern Dutch songs. Since all traditions use the same versification system, metrical forms are inherently comparable, which makes cross-cultural study possible. At the same time, all data comes from corpora of different designs and principles (Appendix 1), and even includes different textual domains (Dutch songs). The heterogeneity of sources ensures that our findings would not result from some artifact of corpus construction.   We devise a language-independent methodology to represent each poem as a set of abstracted semantic features (latent topics inferred from all texts within a corpus) and rely on aggregation and random sampling to access meter-meaning relationships.

The main question of this study is whether poems written in a particular meter also tend to employ similar topics \textbf{(H1)}. We formalize this as clustering and classification tasks. Do poems written in one meter tend to group into coherent semantic clusters? And can a meter be recognized based on knowledge of nothing but the abstracted semantic features of corresponding poems? 

These queries, in turn, raise the issue of semantic diffusion and retention. It was informally suggested that the semantic halo may become more diffuse in modern traditions as time goes on and more poems are written by an expanding pool of authors \cite{gasparov_metr_1999}. We expect clustering to be less efficient in the later phases of a tradition compared to its early stages when we use the same sets of meters and similar sampling strategies  \textbf{(H2)}. This should be evident in all the corpora except for the Dutch songs, which come from the early modern period and represent a tradition grounded in oral performance and genre continuity. This tradition is presumably less open to sudden innovation. 

Finally, we expect that despite the diffusion of the semantic halo, poetic traditions retain a historical connection to the halo's earlier states. If this is true, then it should be possible to recognize meters from earlier periods using models only exposed  to later data and vice versa \textbf{(H3)}.

To test these assumptions, we use the abstracted semantic features of poems (represented as topic probabilities) to recognize their metrical organization (represented as the unambiguous labels of metrical types). The distribution of topics or co-occurring groups of words in each poem is inferred with the aid of a generative Latent Dirichlet Allocation (LDA) model \cite{blei_latent_2003} trained for each lexically simplified corpus. Metrical labeling of poems combines information about the general metrical scheme used (e.g. iamb) and the particular type of this scheme (e.g. pentameter), as shown in Table \ref{tab1}. We understand  metrical types to be the main verse forms  reproduced within and across traditions; they each have a distinct historical lineage that may be reconstructed to a greater or lesser extent. To take one example, English iambic pentameter, as introduced in the 14th century, may be traced to 10-syllable French isosyllabic meter, which developed from the Italian 11-syllable form. On the other hand, the so-called ``ballad'' meter (Iamb 4-3, ``common measure'') originated in local Anglo-Saxon accentual versification and had a distinct range of folksong associations \cite{saintsbury_history_1995, tarlinskaja_english_1976, gasparov_history_1996}. Where possible, the poems in our corpora were individually annotated with a single unambiguous label related to their metrical type. (The limitations of this approach are discussed in Appendix 2.)

\begin{table}[t!]
  \begin{tabular}{cclll}
    \toprule
    Meter & Foot & Pattern & Metrical Type & Label \\
    \midrule
    Iamb  & WS & WS$|$WS$|$WS$|$WS$|$WS & iambic pentameter & I5 \\
    & & Thus was$|$I, slee$|$ping, by $|$ a bro$|$ther's hand & \\
    & & Of life,$|$ of crown,$|$ of queen,$|$ at once$|$ dispatch'd & \\
    \midrule
    Trochee & SW & SW$|$SW$|$SW$|$S(W) & trochaic tetrameter & T4 \\
    & & Tell me$|$   not in$|$  mournful $|$ numbers, & \\
    & & Life is $|$  but  an $|$  empty $|$ dream & \\
    \midrule
    Dactyl  & SWW & SWW$|$SWW$|$SWW$|$S(WW)  & dactylic tetrameter & D4 \\
     & & Brightest and$|$  best of the $|$ sons of the$|$ morning   & \\
    \midrule
    Amphibrach & WSW & WSW$|$WSW$|$WSW$|$WS(W)  & amphibrachic tetrameter & A4 \\ 
     & & Oh, hush thee,$|$ my baby$|$, thy sire was $|$ a knight   & \\
     & & Thy mother $|$ a lady$|$ both lovely $|$ and bright   & \\
    \midrule
    Anapest & WWS & WWS$|$WWS & anapestic dimeter & An2 \\
     & & He is gone $|$ on the moun$|$tain     & \\
     & & He is lost $|$ to the for$|$est & \\
  \bottomrule
\end{tabular}
  \caption{Examples of metrical types and labeling strategies. \textbf{S} - denotes a strong position in the foot (stress expected), \textbf{W} - weak.}
  \label{tab1}
\end{table}

We assume that meter \& meaning association to be impossible to trace at a single-poem level since the semantic halo is not a deterministic mechanism \cite{gasparov_metr_1999} that  prescribes meanings to texts. Rather this mechanism is probabilistic and observable at the level of central tendencies, (which might, in turn, depend on the aesthetic conventions of a group or a tradition). Given these factors and  the skewed distribution of the metrical forms used in a tradition (see Supplementary Fig.~\ref{S3_Fig}), we rely on an approximation of each meter's semantic features based on random equal-sized samples of poems from a set of the most frequently used forms in the tradition.

Our analysis confirms the presence of the semantic halo of meter in all observed European traditions. In all cases except for the Dutch example, we also observe the predicted historical decline in the strength of this association. When the sample size is large enough,the cross-period classification exceeds the random baseline in all cases. There are, however, significant differences among the corpora that may reflect different levels of exposure to change in the  poetic traditions.


\section*{Results}


\subsection*{Meter and meaning (H1)}

We approach the association between meters and topics as a clustering experiment: we expect that independent samples of poems written in the same meter, as represented by vectors of topic probabilities, will tend to group together.  We limit the data to meters that are sufficiently common  (\# of poems $>$ 500) and then divide them into random samples of 100 poems per meter and aggregate topic probabilities within each sample. $K$-means clustering is then used to group the samples into $k$ clusters, with $k$ being set as the number of distinct meters. The resulting clusters are  compared against the true labels, i.e. groups based on the actual meters.

\begin{figure}[!t]
    \centering
    \begin{subfigure}[b]{0.05\textwidth}
        \centering
        \caption{}
        \label{fig:ari_cp}
        \vspace{5cm}
    \end{subfigure}    
    \begin{subfigure}[b]{0.4\textwidth}
        \centering
        \includegraphics[width=\textwidth]{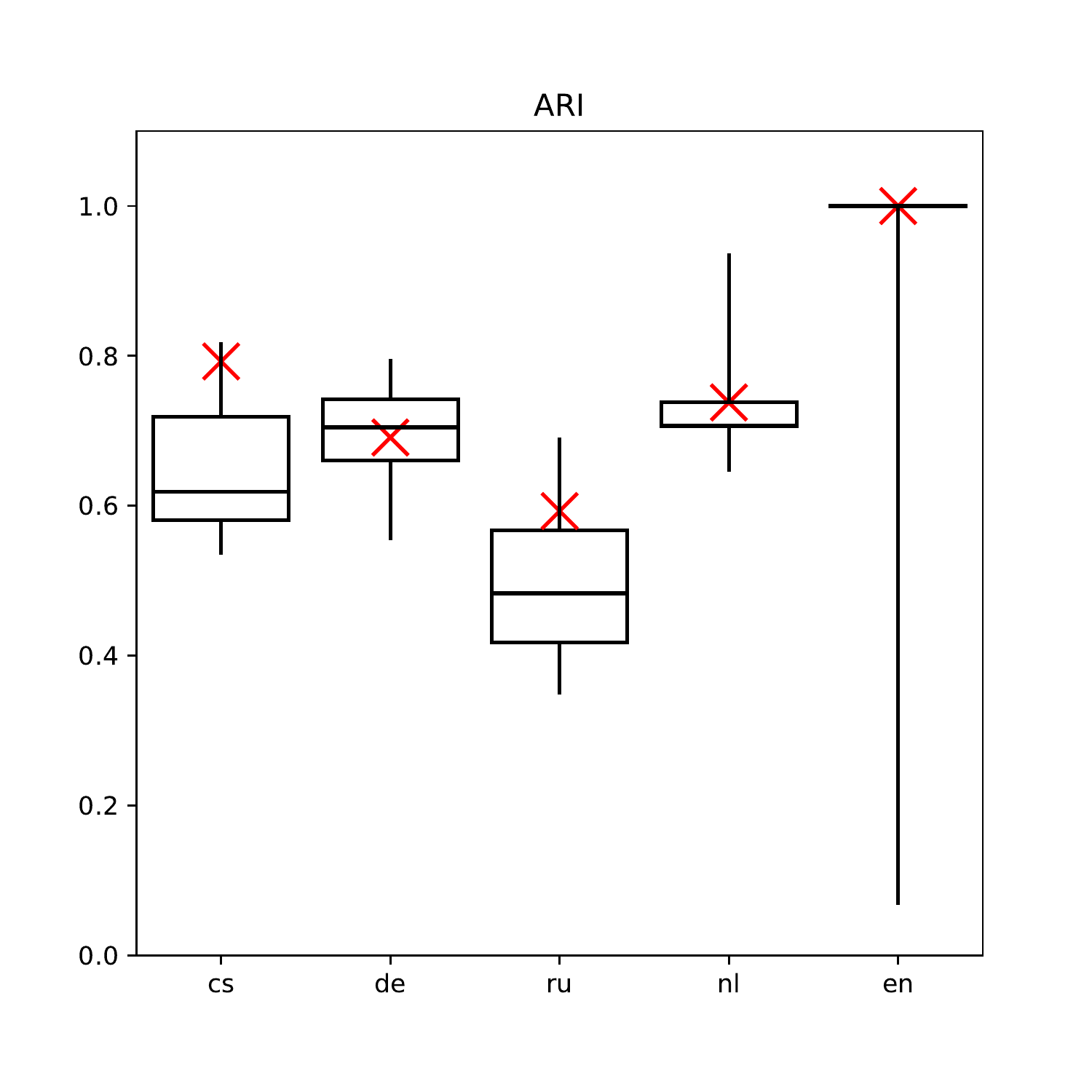}
        \vspace{-10mm}
    \end{subfigure}  
    \hspace{1em}
    \begin{subfigure}[b]{0.05\textwidth}
        \centering
        \caption{}
        \label{fig:meters_cs}
        \vspace{5cm}
    \end{subfigure}
    \begin{subfigure}[b]{0.4\textwidth}
        \centering
        \includegraphics[width=\textwidth]{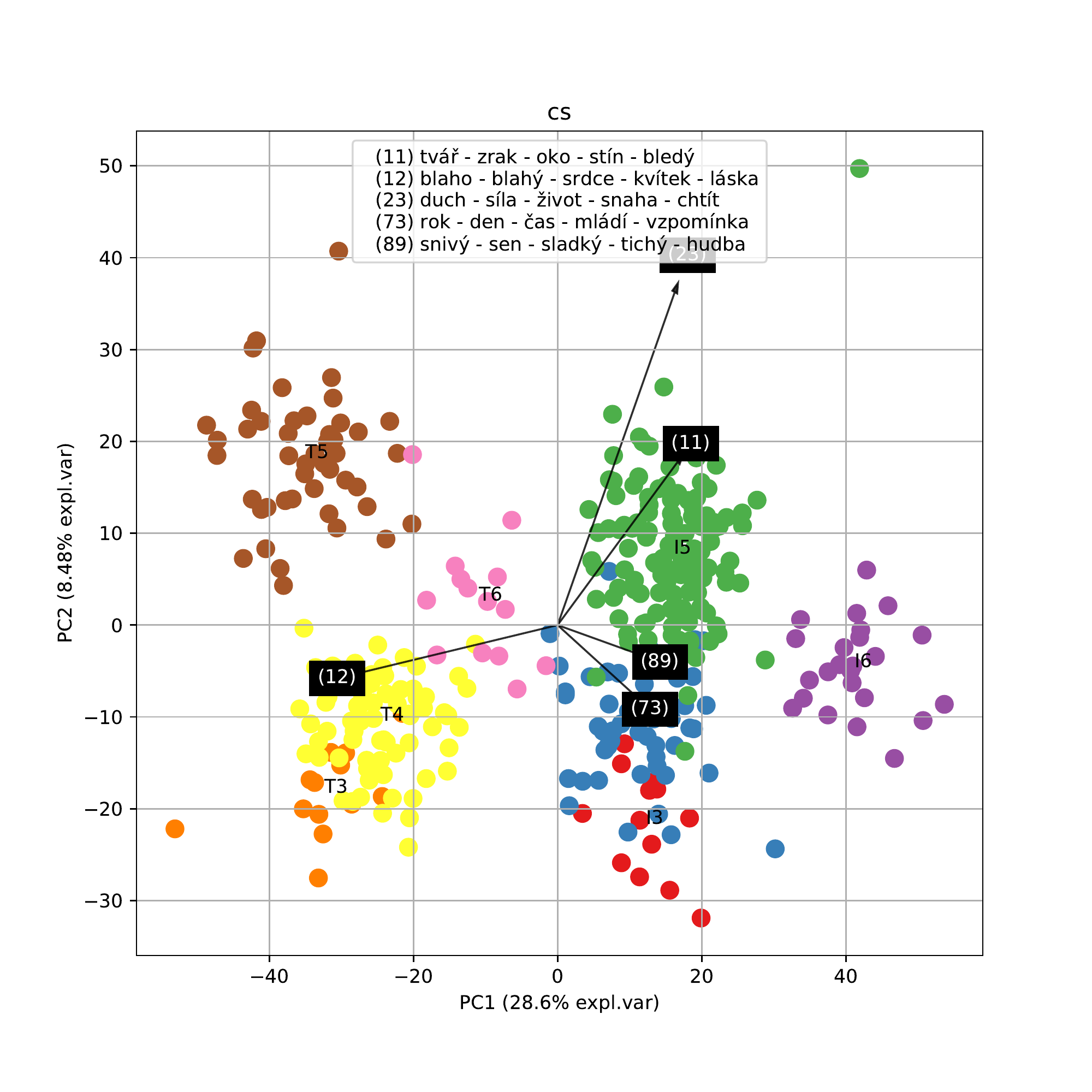}
        \vspace{-10mm}
    \end{subfigure}\\
    \begin{subfigure}[b]{0.05\textwidth}
        \centering
        \caption{}
        \label{fig:meters_de}
        \vspace{5cm}
    \end{subfigure}
    \begin{subfigure}[b]{0.4\textwidth}
        \centering
        \includegraphics[width=\textwidth]{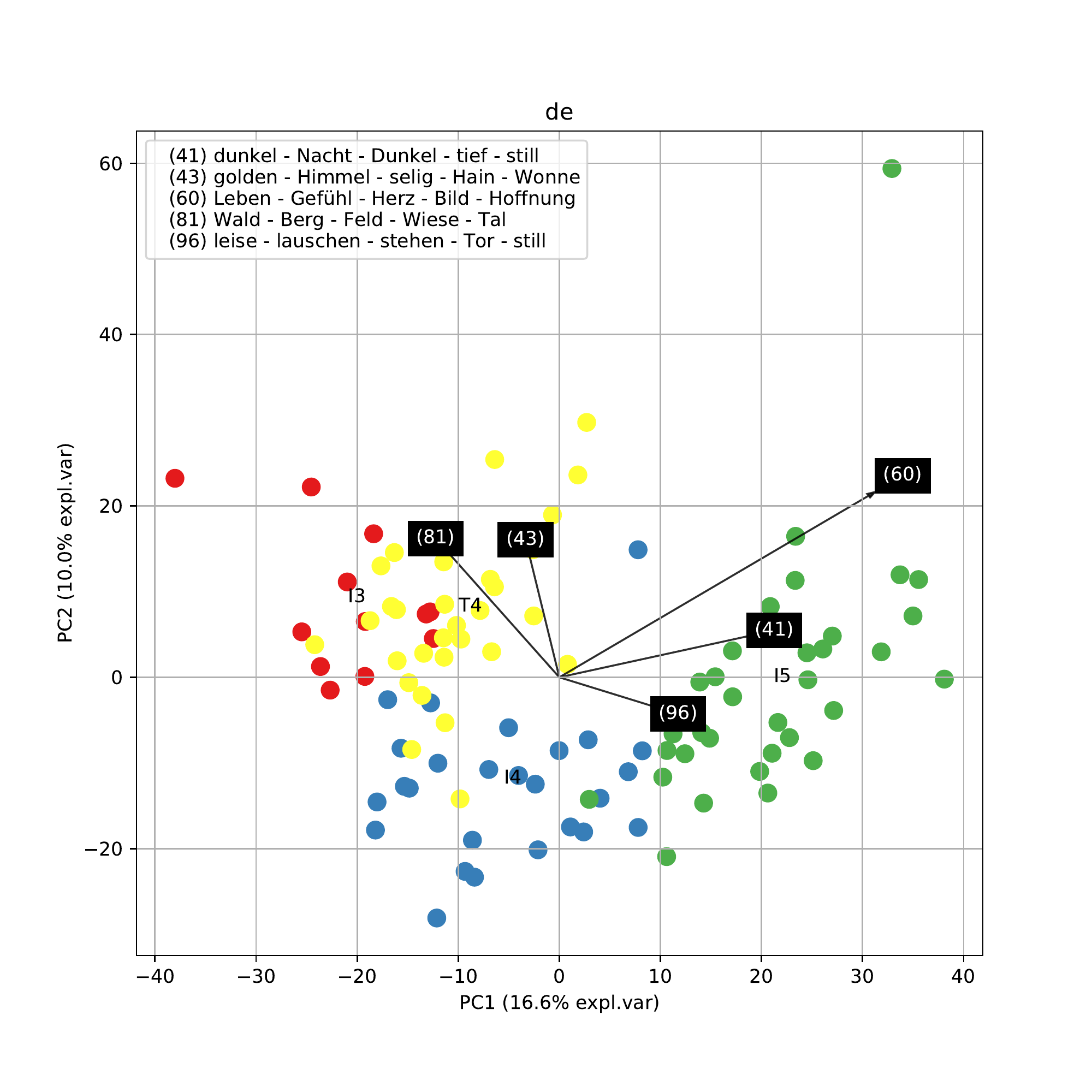}
        \vspace{-10mm}
    \end{subfigure}
     \hspace{1em}
    \begin{subfigure}[b]{0.05\textwidth}
        \centering
        \caption{}
        \label{fig:meters_ru}
        \vspace{5cm}
    \end{subfigure}
    \begin{subfigure}[b]{0.4\textwidth}
        \centering
        \includegraphics[width=\textwidth]{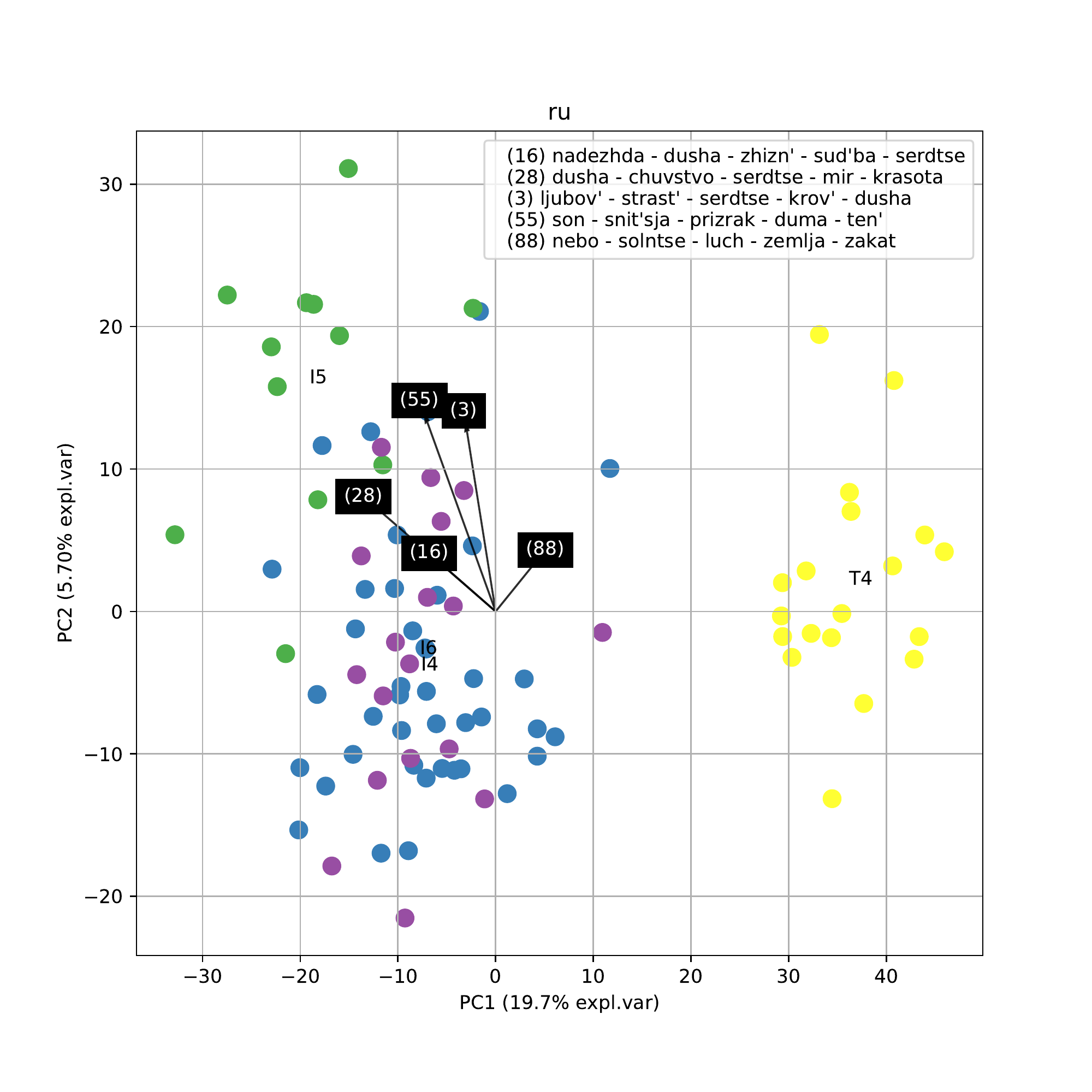}
        \vspace{-10mm}
    \end{subfigure}\\
    \begin{subfigure}[b]{0.05\textwidth}
        \centering
        \caption{}
        \label{fig:meters_nl}
        \vspace{5.5cm}
    \end{subfigure}
    \begin{subfigure}[b]{0.4\textwidth}
        \centering
        \includegraphics[width=\textwidth]{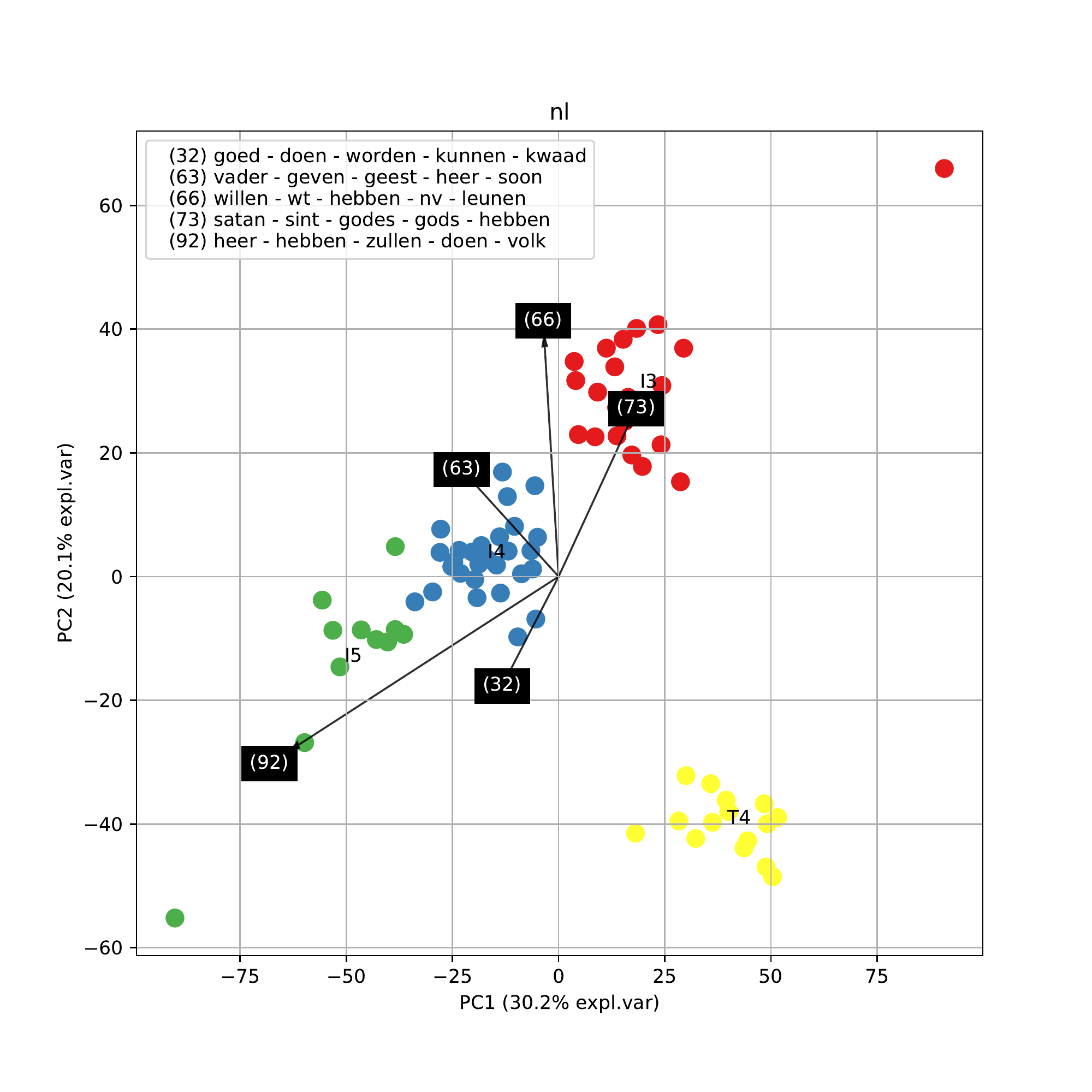}
    \end{subfigure}
    \hspace{1em}
    \begin{subfigure}[b]{0.05\textwidth}
        \centering
        \caption{}
        \label{fig:meters_en}
        \vspace{5.5cm}
    \end{subfigure}
    \begin{subfigure}[b]{0.4\textwidth}
        \centering
        \includegraphics[width=\textwidth]{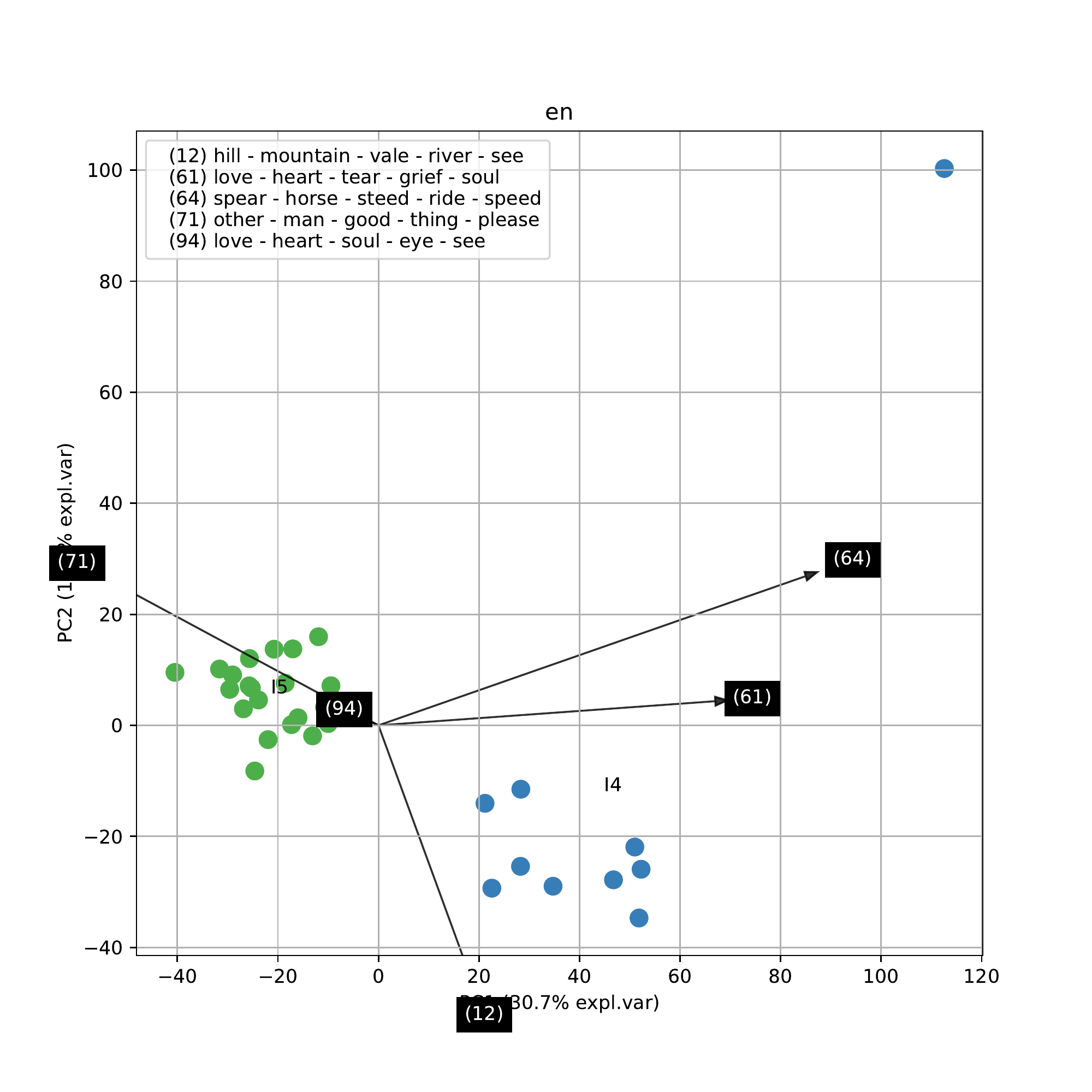}
    \end{subfigure}  
    \begin{subfigure}[b]{1\textwidth}
        \centering
        \includegraphics[width=\textwidth]{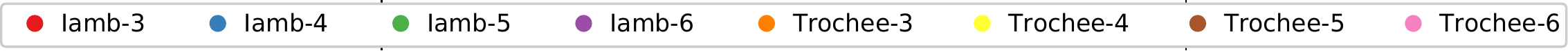}
    \end{subfigure}          
    \caption{Random 100-poem samples taken without replacement per meter in vector spaces defined by LDA topic models.
    \subref{fig:ari_cp}, Adjusted Rand index of $k$-means clustering  (whiskers give the 5th- to 95th-percentile range). 10,000 random samplings. Crosses give the ARI of the samplings presented in PCA biblots.
    \subref{fig:meters_cs}--\subref{fig:meters_en}, PCA biplots of Czech (8 meters), German (4 meters), Russian (4 meters), Dutch (4 meters) and English data (2 meters) respectively with eigenvectors for the 5 most contributing topics. Single random sampling.}
    \label{fig:meters}
\end{figure}

The procedure is repeated 10,000 times with each corpus (Fig.~\ref{fig:ari_cp}). All results point to a high level of association between metrical forms and the semantic features of corresponding poems in each of the observed corpora. This provides strong support for the presence of the semantic halo across all traditions and confirms H1. To illustrate the extent of the association between form and meaning, one such random sample from each corpus is represented by means of two-dimensional PCA biplots (Fig.~\ref{fig:meters_cs}--\ref{fig:meters_en}).

An additional supervised classification experiment (a support vector machine) corroborates the evidence with even smaller samples (Fig.~\ref{fig:svm}: first boxplot series). 

\subsection*{Diffusion over time (H2)}

We expect the semantics of meter to become less recognizable over time: more and more poems emerged within aesthetics that reportedly shifted from normative (i.e. generic conventions in an example-based literature) to individualistic (i.e. based on innovation, inspiration and organistic models of art \cite{abrams_mirror_1953}). This hypothesis is, however, generally limited to canonical 19th-century poetry, and there is no reason to apply it to early modern popular songs (i.e. our Dutch corpus). As a tradition which relied on oral performance and the persistence of popular melodies, these songs are expected to maintain more stable generic lineages.

To test the hypothesis, we split each corpus into two subcorpora based on the date of publication of particular poems. (Both the amount of data at our disposal and the distribution of meters over time prevent us from dividing the corpora according to a more granular chronology or partitioning the English data in a similar way.) In each subcorpus, we test the strength of the association between metrical labels and sample-wide aggregated semantic features in the manner described above. To ensure comparability, we use the same set of meters and draw a fixed number of samples from the two parts of each corpus.

Based on H2, we expect to see a decrease in clustering accuracy in the later stage of the timeline when compared to the earlier part. We observe this trend in the Czech, German and Russian corpora but not in the the Dutch songs where forms and their semantics maintain similar levels of association over 200 years (Table~\ref{tab2}).

\begin{table}[t!]
\centering
\begin{tabular}{cccccc}
\hline
\multirow{2}{*}{Language} & \multirow{2}{*}{Time span} & \multicolumn{2}{c}{ARI} & \multirow{2}{*}{Meters} & \multirow{2}{*}{\# of samples per meter}\\
 & & mean & std. dev. & & \\
\hline

\multirow{2}{*}{Czech}
& 1800--1859 & 0.994 & 0.034 & \multirow{2}{*}{I5, T4, T5} & \multirow{2}{*}{5}\\
& 1860--1919 & 0.878 & 0.152 & & \\
\multirow{2}{*}{German}
& 1750--1824 & 0.958 & 0.098 & \multirow{2}{*}{I4, I5, T4} & \multirow{2}{*}{5}\\
& 1825--1900 & 0.444 & 0.226 & & \\
\multirow{2}{*}{Russian}
& 1800--1859 & 0.715 & 0.181 & \multirow{2}{*}{I4, I5, T4} & \multirow{2}{*}{5}\\
& 1860--1899 & 0.642 & 0.218 & & \\
\multirow{2}{*}{Dutch}
& 1550--1649 & 1 & 0 & \multirow{2}{*}{I3, I5, T4} & \multirow{2}{*}{4}\\
& 1650--1750 & 0.995 & 0.013 & & \\
\hline
\end{tabular}
\caption{Adjusted Rand index of $k$-means clustering in different periods (random 100-poem samples). 10,000 iterations.}
\label{tab2}
\end{table}


\subsection*{Retention of forms over time (H3)}

\begin{figure}[!t]
    \centering
    \begin{subfigure}[b]{0.05\textwidth}
        \centering
        \caption{}
        \label{fig:svm_cs}
        \vspace{5cm}
    \end{subfigure}
    \begin{subfigure}[b]{0.4\textwidth}
        \centering
        \includegraphics[width=\textwidth]{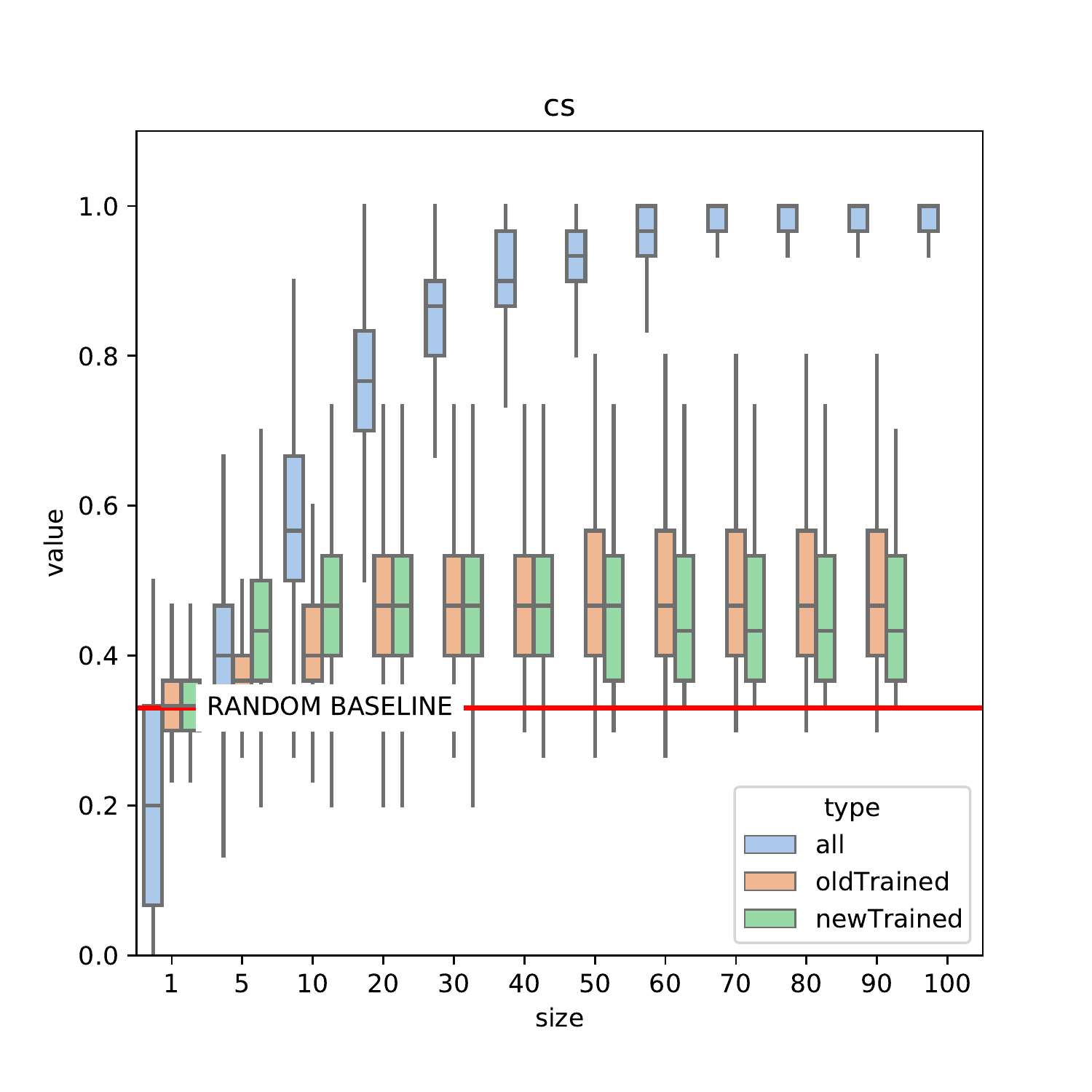}
        \vspace{-10mm}
    \end{subfigure}
    \hspace{1em}
    \begin{subfigure}[b]{0.05\textwidth}
        \centering
        \caption{}
        \label{fig:svm_de}
        \vspace{5cm}
    \end{subfigure}
    \begin{subfigure}[b]{0.4\textwidth}
        \centering
        \includegraphics[width=\textwidth]{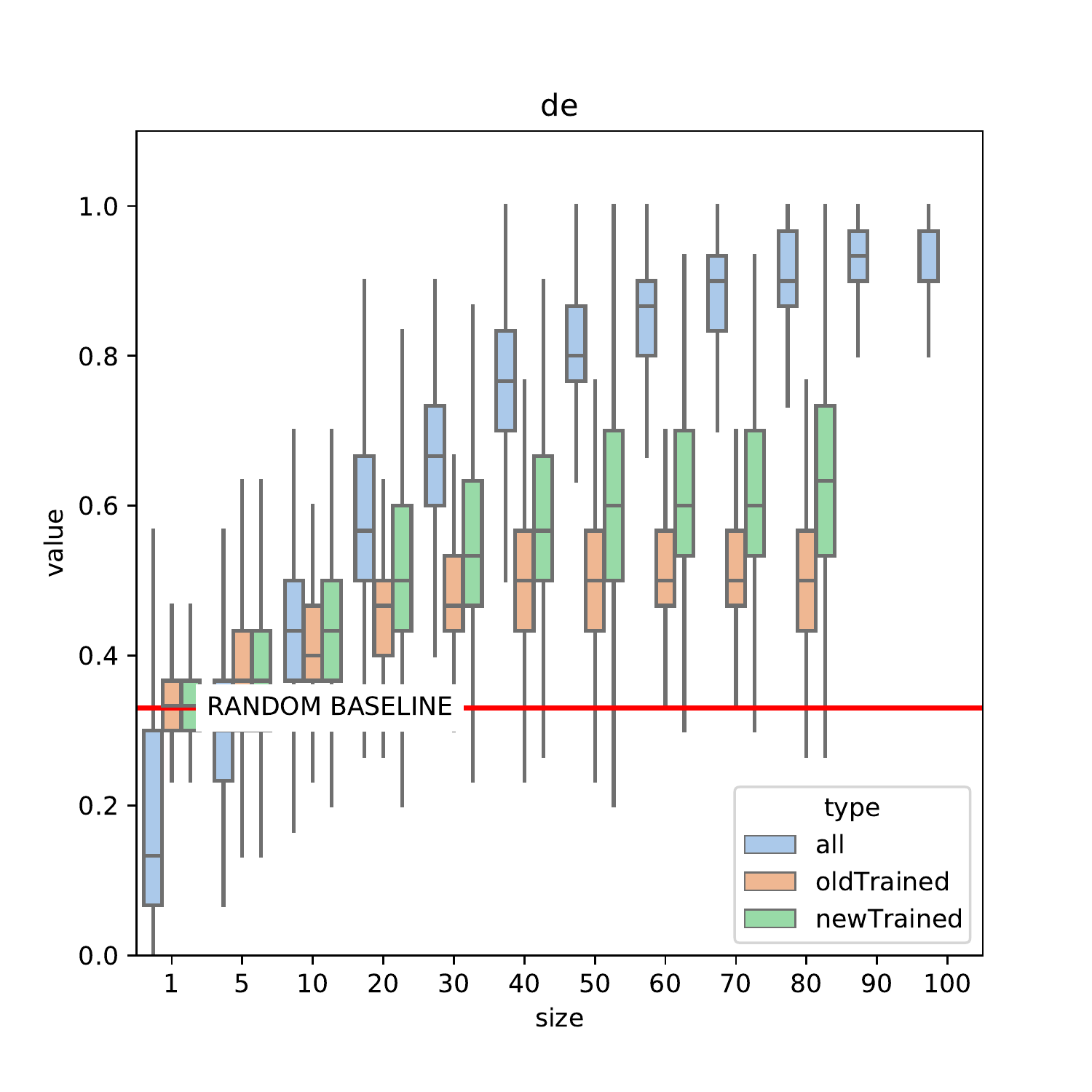}
        \vspace{-10mm}
    \end{subfigure}\\
    \begin{subfigure}[b]{0.05\textwidth}
        \centering
        \caption{}
        \label{fig:svm_ru}
        \vspace{5cm}
    \end{subfigure}
    \begin{subfigure}[b]{0.4\textwidth}
        \centering
        \includegraphics[width=\textwidth]{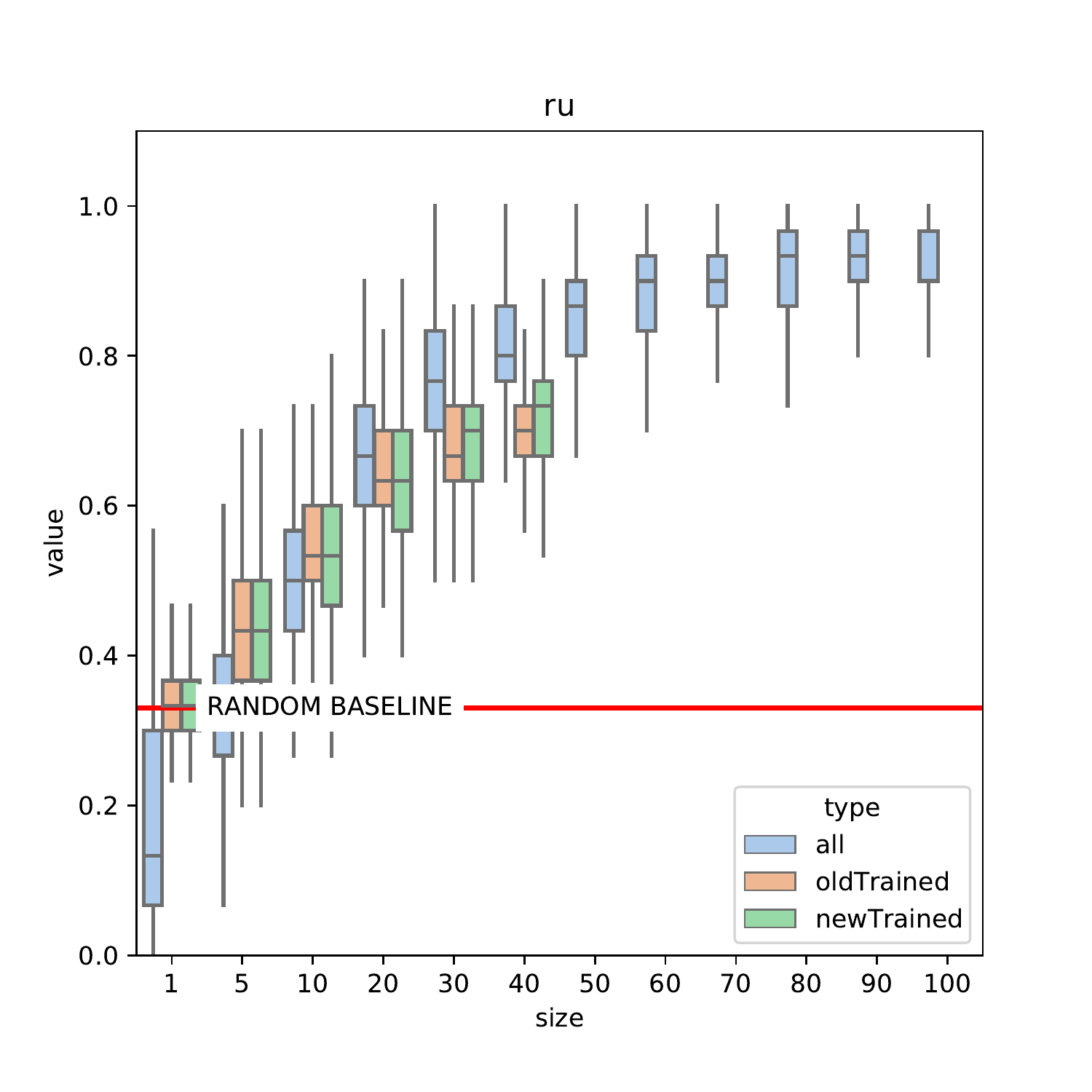}
        \vspace{-10mm}
    \end{subfigure}
    \hspace{1em}
    \begin{subfigure}[b]{0.05\textwidth}
        \centering
        \caption{}
        \label{fig:svm_nl}
        \vspace{5cm}
    \end{subfigure}
    \begin{subfigure}[b]{0.4\textwidth}
        \centering
        \includegraphics[width=\textwidth]{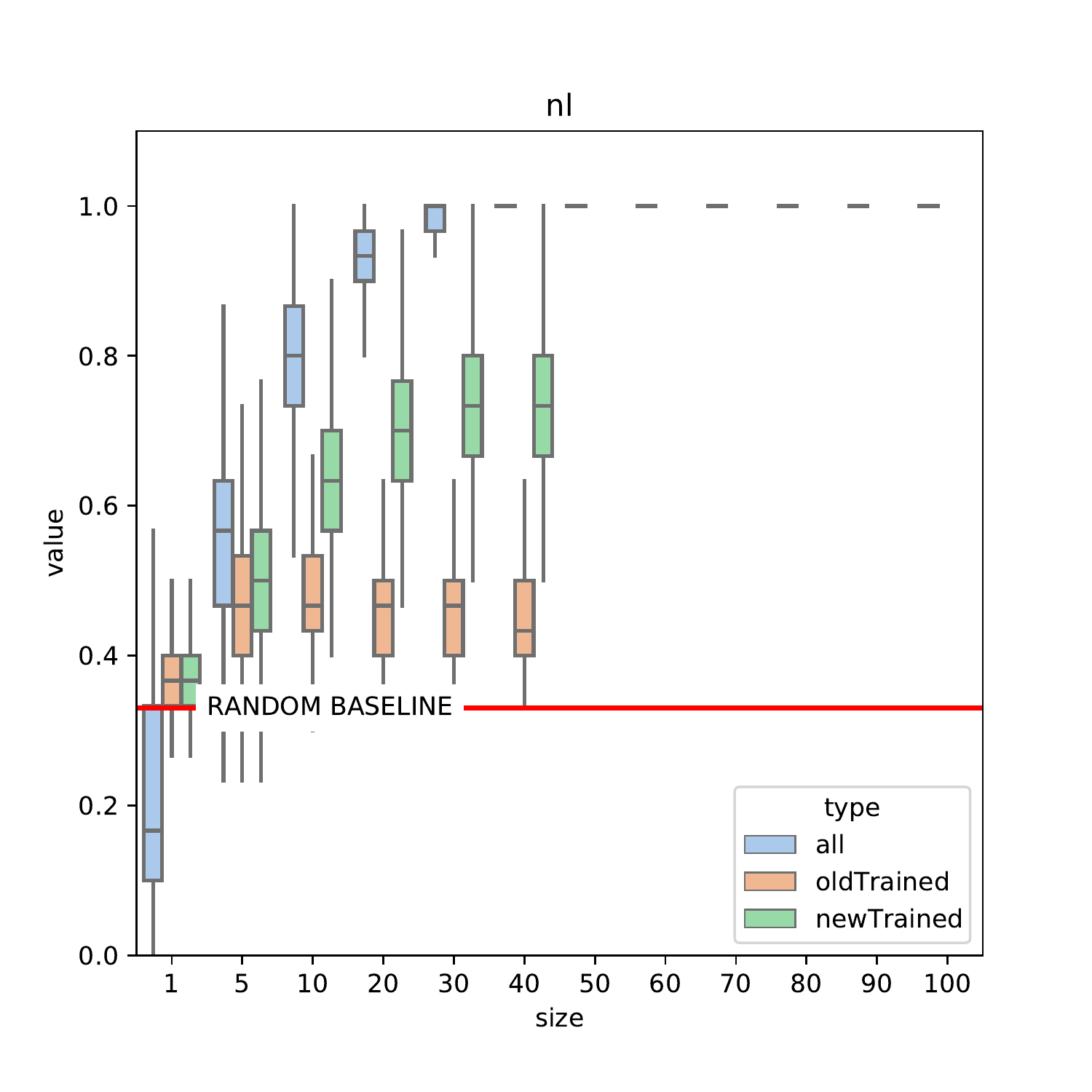}
        \vspace{-10mm}
    \end{subfigure}\\
    \begin{subfigure}[b]{0.05\textwidth}
        \centering
        \caption{}
        \label{fig:svm_en}
        \vspace{5.5cm}
    \end{subfigure}
    \begin{subfigure}[b]{0.4\textwidth}
        \centering
        \includegraphics[width=\textwidth]{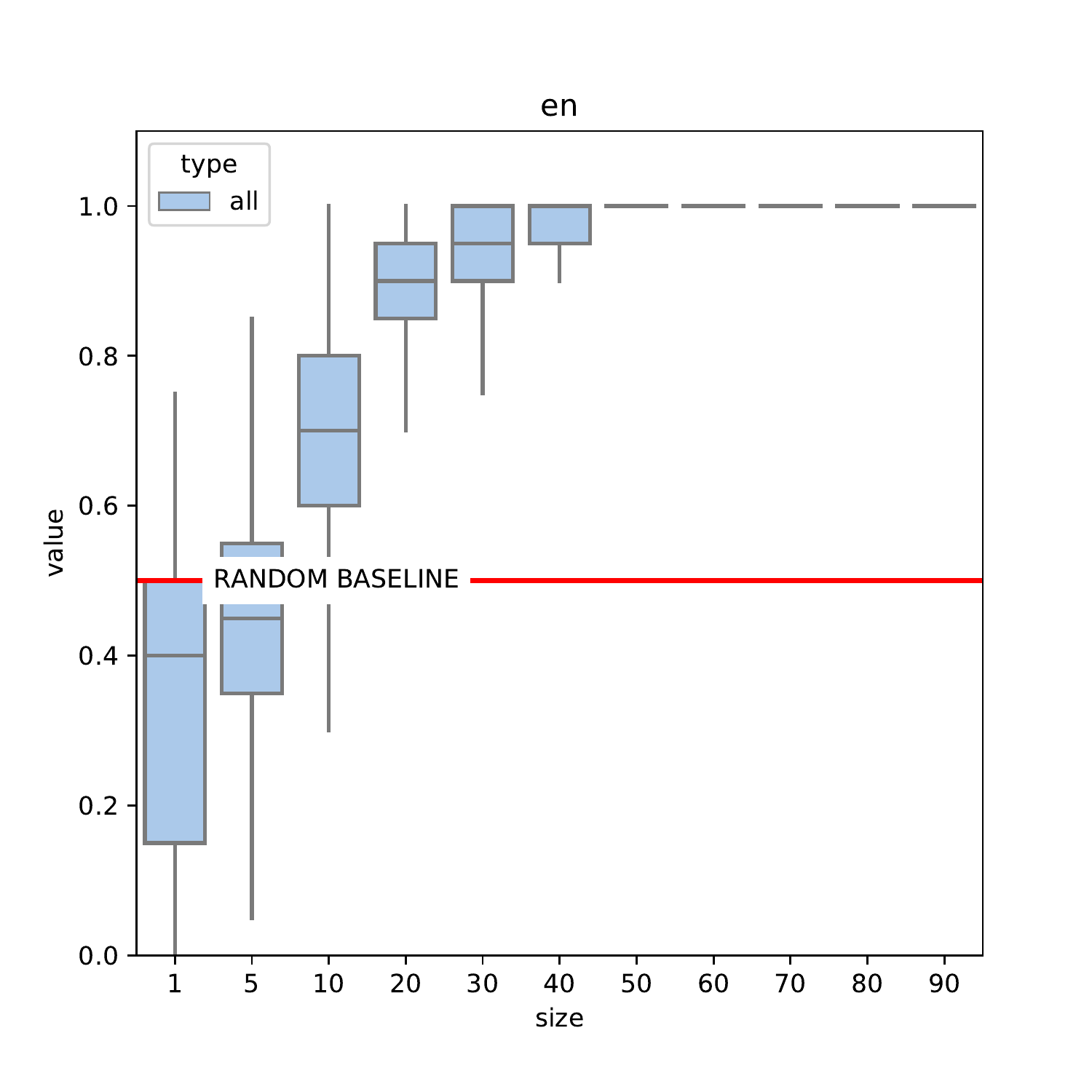}
    \end{subfigure} 
    \hspace{1em}
    \caption{Accuracy of SVM classifications: predicting meter with vectors defined by topic probabilities in random samples of poems (sample size $\in \{1,5\} \cup \{10,20,\ldots, 100\}$). (1) trained and tested on the entire dataset (leave-one-out cross-validation), (2) trained on earlier data and tested on later data, (3) trained on later data and tested on earlier data. 10,000 iterations.}
    \label{fig:svm}
\end{figure}

Despite changes over time in the strength of the semantic halo, we expect to see historical continuity in the use of various forms so that we may predict the later stages of a meter's semantics by knowing its earlier semantic features and vice versa. To test this, we perform supervised classification on each corpus: the training set is restricted to samples from one subcorpus while the test set only includes poems from the other one. 
The results show significant variation (Fig. \ref{fig:svm}) across the corpora, which may reflect differences in their sources and/or historical idiosyncrasies in the usage of meter. The Russian tradition exhibits the most stable use of the main metrical forms over the observed period; there is little difference between past and future forms. In contrast, both the German and Dutch corpora show higher semantic recognizability from the future to the past than in the opposite direction. The assymetry suggests semantic accumulation over time, a pattern where later metrical semantics ``enclose'' \cite{chang_divergence_2020, underwood_can_2021} the early usage of a form but are already too different to be recognizable from the past. The recognizability of Czech forms stays stable at a level barely above the random baseline: it is likely connected to this tradition being in a volatile establishment state and changing its metrical preferences midway through the 19th century (Supplementary Fig. \ref{S4_Fig}).

Overall these results provide weak support for H3, but also highlight the high degree of turn-around and variation in the metrical semantics.

\section*{Discussion} 

Our findings strongly support the association between poetic meter and meaning, providing large-scale evidence for the theory of the semantic halo. At least within certain accentual-syllabic systems, European poetic traditions demonstrate their use of conservative mechanisms to produce and retain meaning. Metrical forms easily attract arbitrary semantic features that are reproduced when the form itself is reproduced. The precise mechanisms of this attachment are yet to be understood but different dynamics may be at play. These could include both forces intrinsic to the texts (e.g. the mnemonic function of a meter\cite{rubin_memory_1995}) and institutional factors (popularity, example-based teaching, the literary canon \cite{gronas_cognitive_2010})  that ensure that the same slowly expanding set of texts is disseminated to generations of poets.

By focusing on semantic generalization over poetic nuance, our approach is able to  demonstrate the overarching tendencies that distinguish meters from one another. As Fig. \ref{fig:meters} and distinctive topics for each meter suggest, the most striking thematic distinction occurs between trochaic and iambic meters. Globally, the trochee is associated with the themes of love and song, oral poetry and national romantic sentiment. This highlights the trochee's historical roots: across Europe, folk songs were associated with the trochee or trochaic rhythm \cite{gasparov_metr_1999} and they often took form of regular trochaic meters in modern versification systems.  Iambic forms, on the other hand, can be distinguished by topics that associate with a high prestige poetic style: introspective reflections, religious and existential themes. Dutch songs demonstrate the difference most clearly: in Dutch, iambic songs cover religious topics and are the products of an educated, written tradition, while trochaic songs maintain vernacular themes (love, joy, work) that tie them directly to their origins in oral culture.

The semantic tradition that separates iambs from trochees should not, however, be seen as universal. The semantic similarity of trochee-4 and iamb-3 in German, for example, points to the similar usage of these two particular meters and their historical origins in the European tradition of ``anacreontics'' \cite{gasparov_metr_1999}: lyrical songs of feasts, wine and love that are associated with the neo-classical tradition. During the Romantic period, anacreontics easily mixed with rediscovered folksongs and vernacular language, which made iamb-3 gravitate towards the trochaic meaning space. The semantic proximity of these two forms is also evident in Russian verse \cite{sela_weak_2020}, which partially inherited the use of this meter from German. Specific cases with a well-documented history are also captured by the model. For example, trochee-5 in Czech  was the meter embraced by elites during the national revival and its distinctive topics reflect national pathos, folk imagery and civil uprising \cite{sgallova2012}. Russian trochee-5, on the other hand, was initially a rare form. A single extremely popular poem in this meter launched a tradition that bears some similarity to the ``founder'' poem to this day: the LDA model is able to recover topics that refer to its semantic halo that scholars summarise as ``introspective travel on the road (real or metaphorical) at night'' \cite{gasparov_metr_1999}. 

Our conclusions about similar meter and meaning association processes in European verse have obvious limitations. First, the three main traditions observed in our study are closely related and did not develop in isolation: German metrical models played a foundational role in establishing modern verse in both Russian and Czech. This means that deeply rooted metrical associations (e.g. the difference between a trochee and a iamb) may result from common origins or cultural proximity, and not from differentiation as a common function of meter. A second more important limit relates to the level of our recognition of poetic forms. In this work, we aggregate different stanzas and rhyme schemes under a few metrical labels, but these forms often had their own distinct usage traditions. In fact, stanza organization may be more relevant than meter for distinguishing genres in languages where verse regularity is based exclusively on the number of syllables (e.g. French, Italian, Spanish). In other words, we cannot define a ``poetic form'' universally for every poetic tradition, and broader comparative research will need to operate at several levels of abstraction.

We are hopeful that our findings will advance the conversation in metrical studies since they provide the first large-scale formal evidence of an association between meter and meaning in Western poetry. They also pave the way for the incorporation of the explicit modeling of historical processes into literary studies: mechanisms that drive semantic halo processes generally remain a mystery and would require well-defined models that link individual interactions to the observed patterns. This turn to cultural evolutionary framework and cultural transmission models could establish common ground with linguistics, anthropology and social sciences for understanding factors behind the changes and continuities in cultural traditions. Identifying the mechanisms that limit cultural transmission \cite{durham_advances_1990} and generate long-term patterns in creative domains is also important if we are to understand the rate of cultural evolution \cite{lambert_pace_2020,perreault_pace_2012} and the shape of cultural phylogenies \cite{gray_language_2000, obrien_evolutionary_2002, barbrook_phylogeny_1998, youngblood_phylogenetic_2020}. If trochaic tetrameters are more semantically similar to their trochaic tetrameter ``ancestors'' than to any other meter, this form alone could be responsible for establishing and maintaining the divergent traditions within a cultural domain or, as we usually call them, genres.


\section*{Materials and methods}


\subsection*{Corpora}

Our research uses five metrically annotated poetry collections, each of which concerns one language tradition: Czech \cite{plechac2015,plechac2016}, Dutch \cite{van_kranenburg_documenting_2019}, English \cite{jacobs2018}, German \cite{bobenhausen2011,bobenhausen2015} and Russian \cite{grishina_poeticheskii_2009, korchagin_poezija_2015}. These collections have disparate sources and vary in size, chronological scope, general composition principles and survivorship bias (the Russian corpus, for example, favors poems that were reprinted in 20th-century scholarly editions). A summary of the corpora before filtering and pre-processing is provided in Table \ref{tab3}. For more details about each collection and a summary statistics, see Supplementary Figures \ref{S1_Fig}, \ref{S2_Fig}, \ref{S3_Fig}, \ref{S4_Fig}.

\begin{table}[t]
\centering
\begin{tabular}{cccc}
\toprule
Language & Texts & Period & Tokens \\
\midrule
Czech & 69,760 & 18--20th c. & 13,100,898 \\
German & 53,608 & 16--20th c. & 10,462,211 \\
Russian & 17,900 & 18--19th c. & 3,329,352 \\
Dutch & 22,297 & 1550--1750 & 6,562,888 \\
English & 6,448 & 16--19th c. & 2,126,436 \\
\bottomrule
\end{tabular}
\caption{\label{tab3}Summary of poetry corpora used}
\end{table}

This work relies mainly on poetic meters to formalize and distinguish verse forms. Meters - the idealized rules of a text's prosodic composition - arise from the stable recurring rhythmic patterns that organize natural languages. In accentual-syllabic (AS) meters and stress-based languages, meters are organized around the distinction between weak and strong stress. Regularity in AS systems is based on the recurring syllable groups whose conventional names derive from Classical Greek versification. A two-syllable unit in a weak-strong sequence is called an iambic foot; a sequence of feet organizes a poetic line (e.g. iambic pentameter). As a group of contextually aligned lines, a poem gives us information about the metrical type it follows. In classical AS systems, heterogeneous metrical types were dominant in poems (e.g. all lines followed the general pattern of iambic pentameter) although the use of alternating types and free form is not uncommon. In our research, we focus on binary metrical types - the most widespread forms of verse organization in European AS traditions.


\subsection*{Meter recognition}

All the collections in our study were metrically pre-annotated on a line-by-line basis using language-specific rule-based algorithms \cite{bobenhausen2015, grishina_poeticheskii_2009, heuser2010, plechac2015} or manual methods \cite{van_kranenburg_documenting_2019}. Since in all five traditions the dominant versification system is accentual-syllabic all prosodic patterns and verse organization principles are comparable. However, the task of formally describing a poem in a collection is not straightforward since ``metrical form'' may be interpreted on several levels (see Appendix 2). 

Where  possible, each poem in a corpus was assigned a single unambiguous metrical label (e.g. ``iamb-4'' (I4), ``trochee-5'' (T5), etc.). This process involved some heuristics to determine labels: in line with existing annotation simplification principles  \cite{gasparov_metr_1999}, we considered a poem to be of a particular metrical type if at least 80\% of its lines conformed to that same pattern. All heterogeneous cases or non-metrical poems were left unmarked. (Our models were built on all of the available texts, but only the labeled ones were used in the analysis).  


\subsection*{Pre-processing}

All corpora were initially filtered by size to exclude poems that were too short or too long so that the document sizes would remain comparable in the LDA model. We also excluded early modern poems from German and English (Supplementary Fig. \ref{S2_Fig}) to maintain a comparable chronological range across the corpora.
Each corpus went through lemmatization and part-of-speech tagging (\texttt{MorphoDiTa}\cite{strakova2014} was used for the Czech corpus while \texttt{TreeTagger}\cite{schmid1994} was employed for all the other corpora). Afterwards, we applied lexical simplification to words outside the 1000-most- frequent list so that the frequency distribution was less sparse (LDA models are susceptible to noise and work better with a reduced vocabulary \cite{Boyd-Graber:Mimno:Newman-2014, schofield-etal-2017-understanding, uglanova_order_2020}). Low frequency words were replaced with one of their more common contextual neighbors if that neighbor appeared in the list of the 10 semantically closest words. Semantic similarities were determined independently for each corpus with word-embedding models that had been trained on the respective collections to capture the specific semantic relationships of poetic language. For the word-embedding models, we used word2vec \cite{NIPS2013_9aa42b31} implementation from the Python \texttt{gensim} framework \cite{rehurek_lrec}.


\subsection*{Semantic features}

Our study of the overarching semantic relationships of metrical forms required some abstract representation of poetic language that would allow us to summarize a poem's ``content'' at a general level. We used LDA topic models \cite{blei_latent_2003} trained for each of the collections on the non-aggregated data. An LDA model infers topics - groups of co-occurring words - from a collection without supervision. Since LDA is a generative algorithm, each poem in a corpus can be uniformly represented by the topic probabilities that generate its distribution of words. In cultural analytics, topic modeling has become a common way of inferring higher-order semantic properties from a collection of texts so that they can be used for further analysis and reasoning \cite{mauch_evolution_2015, barron_individuals_2018,murdock_exploration_2017, lambert_pace_2020}. While it is less efficient when used with shorter texts \cite{li_enhancing_2017}, it has proven to be generally applicable to poetry without any major reported drawbacks \cite{asgari_conrming_2013, navarro-colorado_poetic_2018, sela_weak_2020}. 

Topic modeling, lexical simplification and lemmatization served another important goal in our study: they mitigated the effects of metrical patterns on morphology and sentence structure \cite{gasparov_linguistics_2008, storey_like_2020}. We additionally checked that our results are independent of pure morphology-based clustering  (Supplementary Table \ref{S2_Table}).

While our results are reported based on an LDA model trained with 100 topics,  our findings remained robust when the number of topics was changed (we conducted tests on models with 20, 50, 100 and 150 topics, Supplementary Table \ref{S1_Table}). The choice of model is therefore not particularly important for the study design. On the other hand,increasing the number of topics may increase the human interpretability of the relationships.


\subsection*{Clustering and classification}

Most of the experiments in this study relied on an unsupervised approach to classification. There were two reasons for this: the first was data scarcity since the popularity of the meters in each tradition follows a skewed distribution. The second reason was the risk of over-fitting: in testing associations between meter and meaning, we wished to capture naturally emerging relationships rather than imposing a fixed view about known ``classes'' on the data.

We used supervised classification solely to model chronological perspectives on meter recognizability (H3). In this case, the training of the model was stratified by time so that learned classes only incorporated ``past'' or ``future'' knowledge about the meter usage in a tradition. 

For $k$-means clustering and the Adjusted Rand Index \cite{hubert_comparing_1985} calculation, we used the implementation provided by the Python library \texttt{scikit-learn} \cite{scikit-learn} (\texttt{sklearn.cluster.KMeans}). The number of clusters was set as the number of distinct meters found in the dataset. 

To test H3, Support Vector Machine classification was also performed with the \texttt{scikit-learn} library, (\texttt{sklearn.SVM.SVC}). We used the classifier with a degree-3 polynomial kernel.


\section*{Code and data availability}

The analysis pipeline, models and input data are available at Zenodo (doi: 10.5281/zenodo.4926549). For the open access corpora (Czech, English, Dutch), we provide each poem as a list of the lemmata it contains. For the proprietary corpora (German, Russian), we provide only topic probabilities for particular poems.


\section*{Supplementary information}


\subsection*{S1 Appendix}
\label{S1_Appendix}
{\bf Corpora details.}
\begin{itemize}
    
\item The Czech data come from the Corpus of Czech Verse created by Petr Plecháč and Robert Kolár.\cite{plechac2015,plechac2016} The entire dataset is available at \url{https://github.com/versotym/corpusCzechVerse}.

\item The German data come from the Metricalizer corpus created by Klemens Bobenhausen and Benjamin Hammerich (\url{https://metricalizer.de/})\cite{bobenhausen2011,bobenhausen2015}. This dataset is proprietary and was kindly provided by its creators for this study.

\item The Russian corpus is part of the Russian National Corpus which is available to researchers upon request \cite{grishina_poeticheskii_2009, korchagin_poezija_2015}. The corpus is metrically pre-annotated but our study uses an improved rhythm recognition algorithm developed by Yurii Zelenkov \cite{plechacsela2021}.

\item English texts come from the Gutenberg English Poetry Corpus compiled by Arthur M. Jacobs.\cite{jacobs2018} Metrical annotation was performed using the Python package \texttt{Prosodic} developed by Ryan Heuser (\url{https://github.com/quadrismegistus/prosodic/}).

\item The early modern Dutch songs used in this study are part of the Dutch Song Database (\url{www.liederenbank.nl}) compiled and hosted by the Meertens Institute in Amsterdam. The database contains more than 175,000 songs in Dutch or Flemish  that date from the Middle Ages through to the twentieth century. The genres include love songs, satirical and  religious works and children's songs. The main sources are songbooks, song sheets (broadsides), song manuscripts and field recordings.

The Czech, German and Russian corpora are the main focus of this study: the poems in these collections cover a comparable cultural niche (prestige poetry) and time span. The Dutch texts come from early modern printed song collections of various sources  and, thus, reflect a specific strand of poetic textual production and circulation. The English collection, on the other hand, contains works scattered over a significant time frame;  particular periods are represented by few texts, and there is little metrical variation. As a result, we use the English \& Dutch collections only as secondary sources. They show the general validity of our claims for material with substantially different structures and origins.

\end{itemize}


\clearpage
\renewcommand{\figurename}{Supplementary Figure}
\renewcommand{\tablename}{Supplementary Table}
\setcounter{figure}{0}  
\setcounter{table}{0}  

\begin{figure}[!t]
    \centering
    \includegraphics[width=\textwidth]{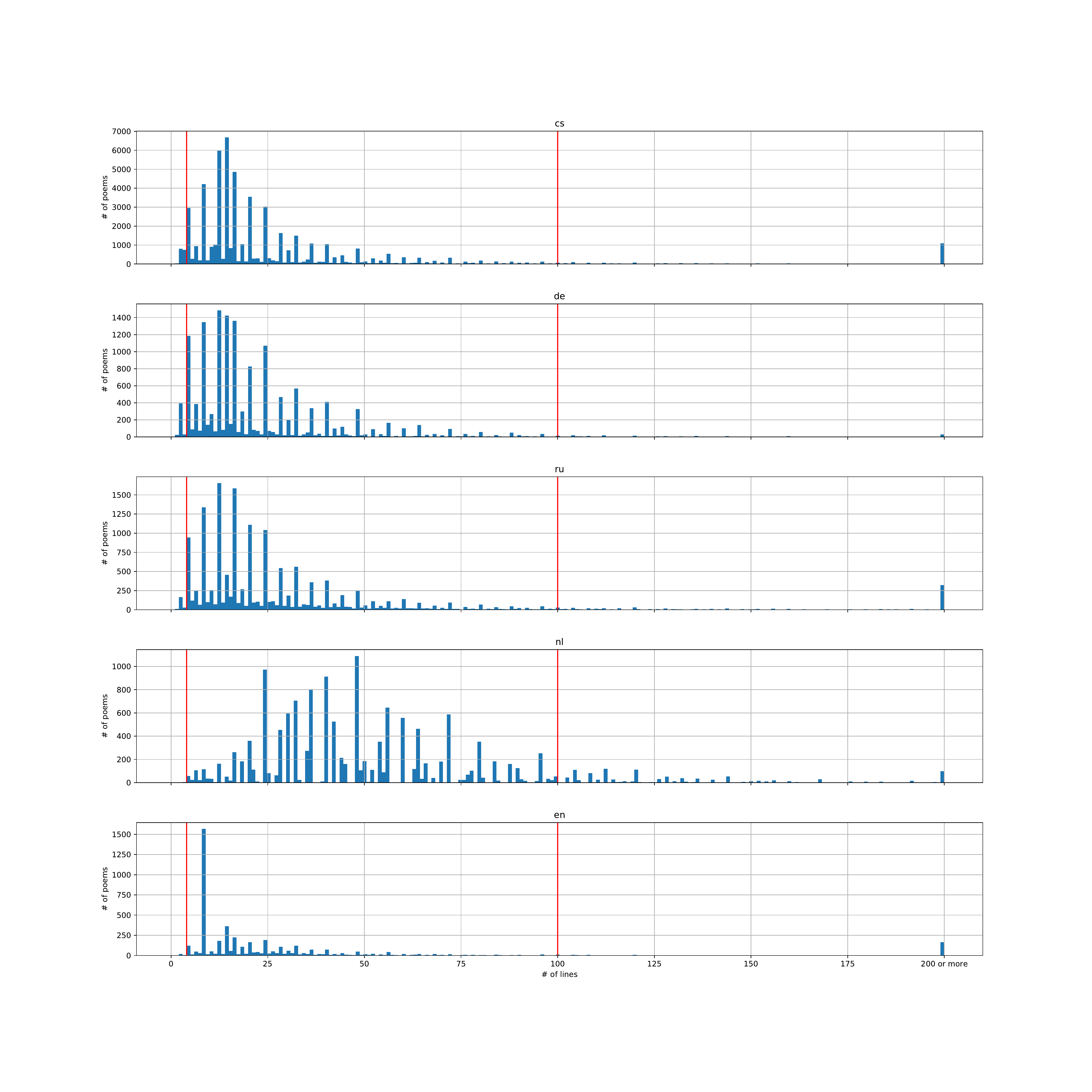}
    \caption{Distribution of poem sizes; red vertical lines mark our filtering cutoffs}
    \label{S1_Fig}
\end{figure}

\begin{figure}
    \centering
    \includegraphics[width=\textwidth]{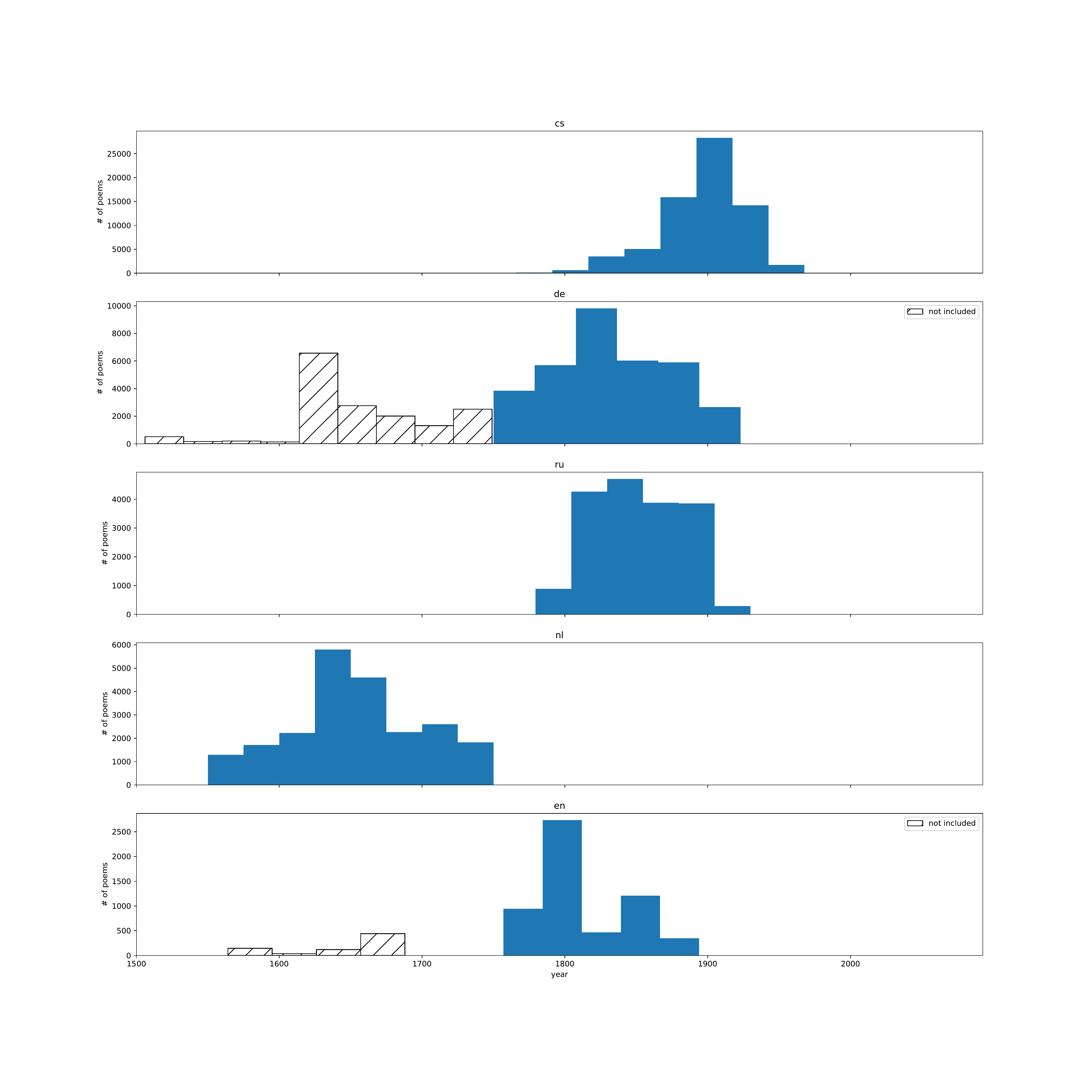}
    \caption{Chronological distribution of poems}
    \label{S2_Fig}
\end{figure}

\begin{figure}
    \centering
    \includegraphics[width=\textwidth]{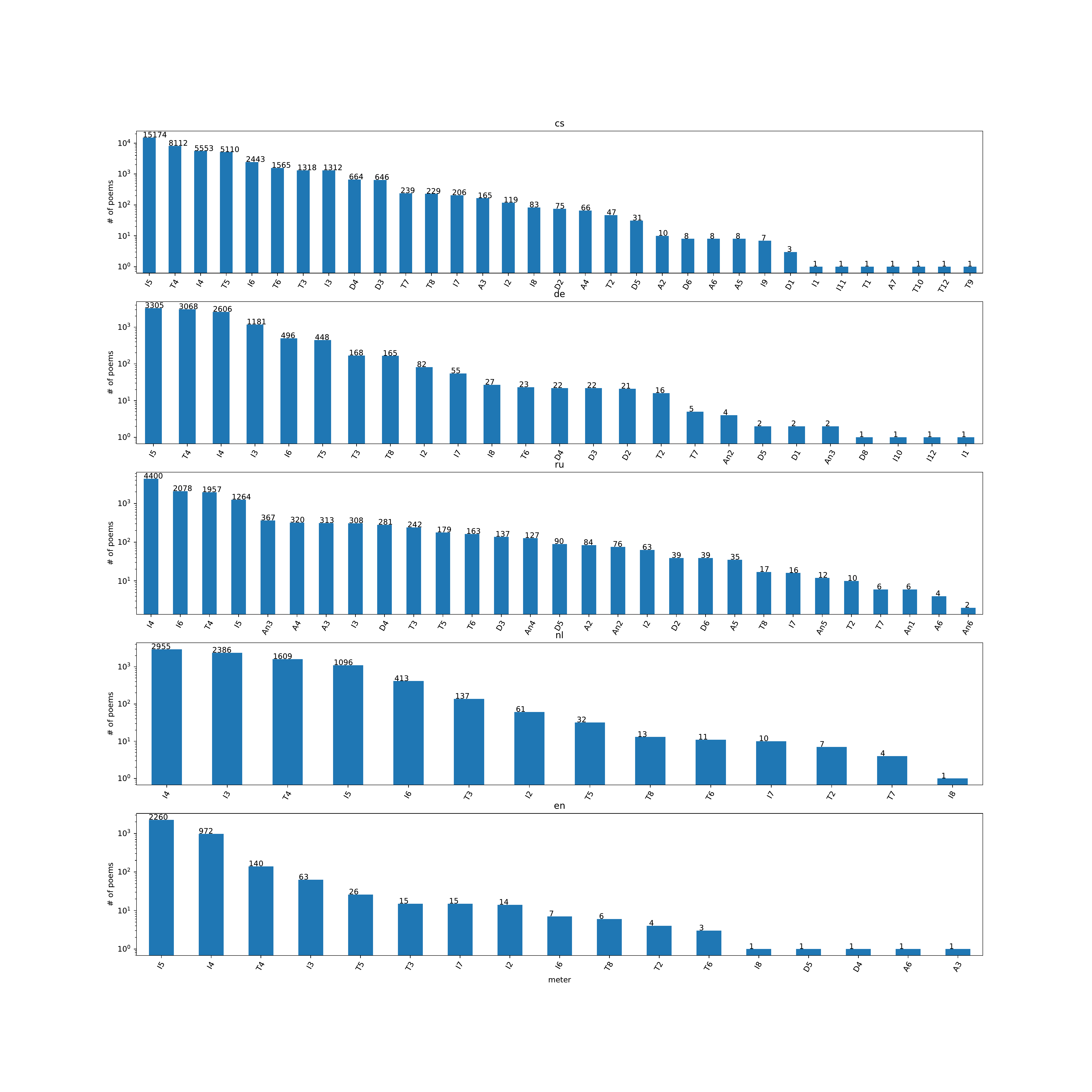}
    \caption{Distribution of metrical types per tradition}
    \label{S3_Fig}
\end{figure}

\begin{figure}
    \centering
    \begin{subfigure}[b]{0.45\textwidth}
        \centering
        \includegraphics[width=\textwidth]{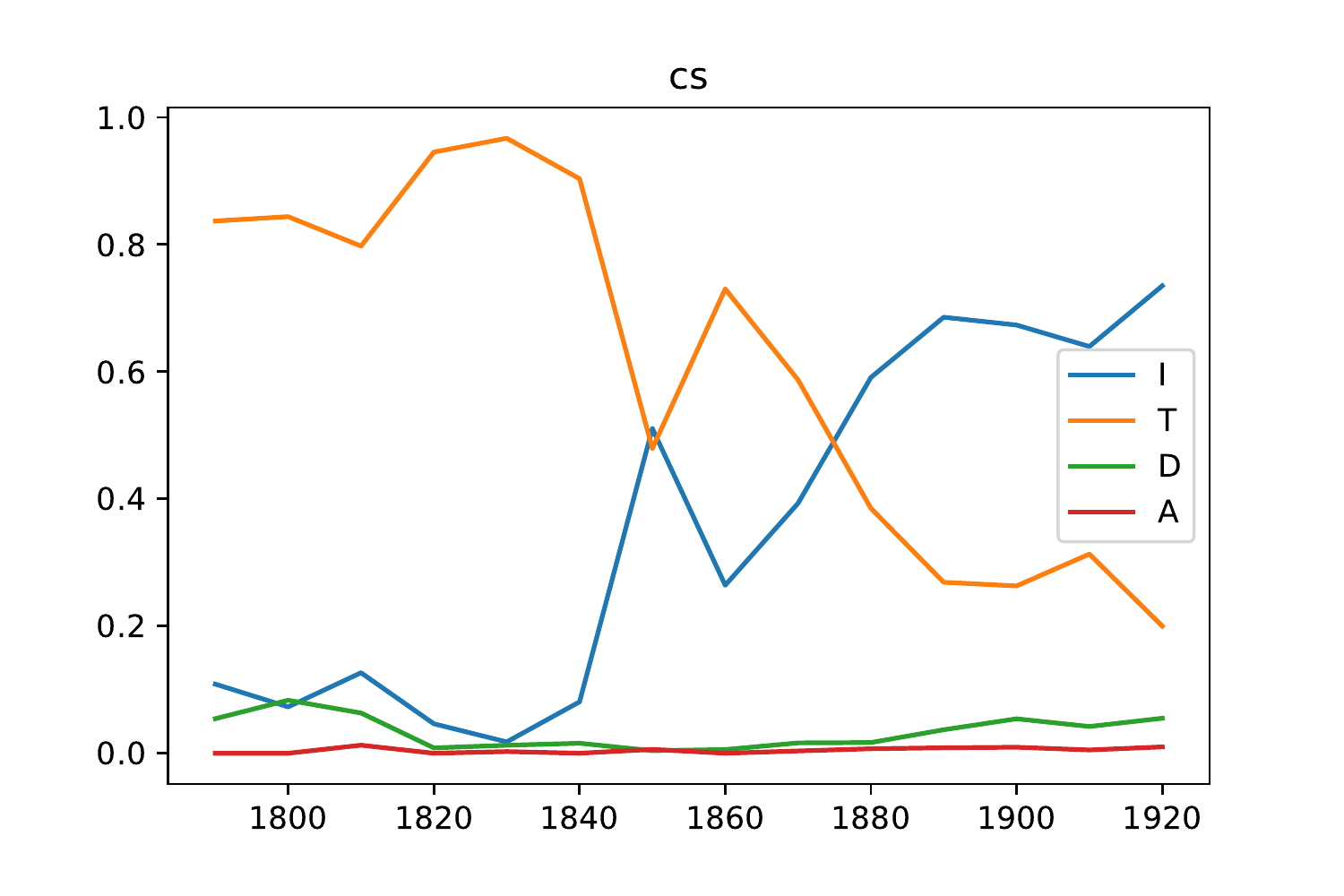}
        \vspace{-10mm}
    \end{subfigure}
    \hspace{1em}
    \begin{subfigure}[b]{0.45\textwidth}
        \centering
        \includegraphics[width=\textwidth]{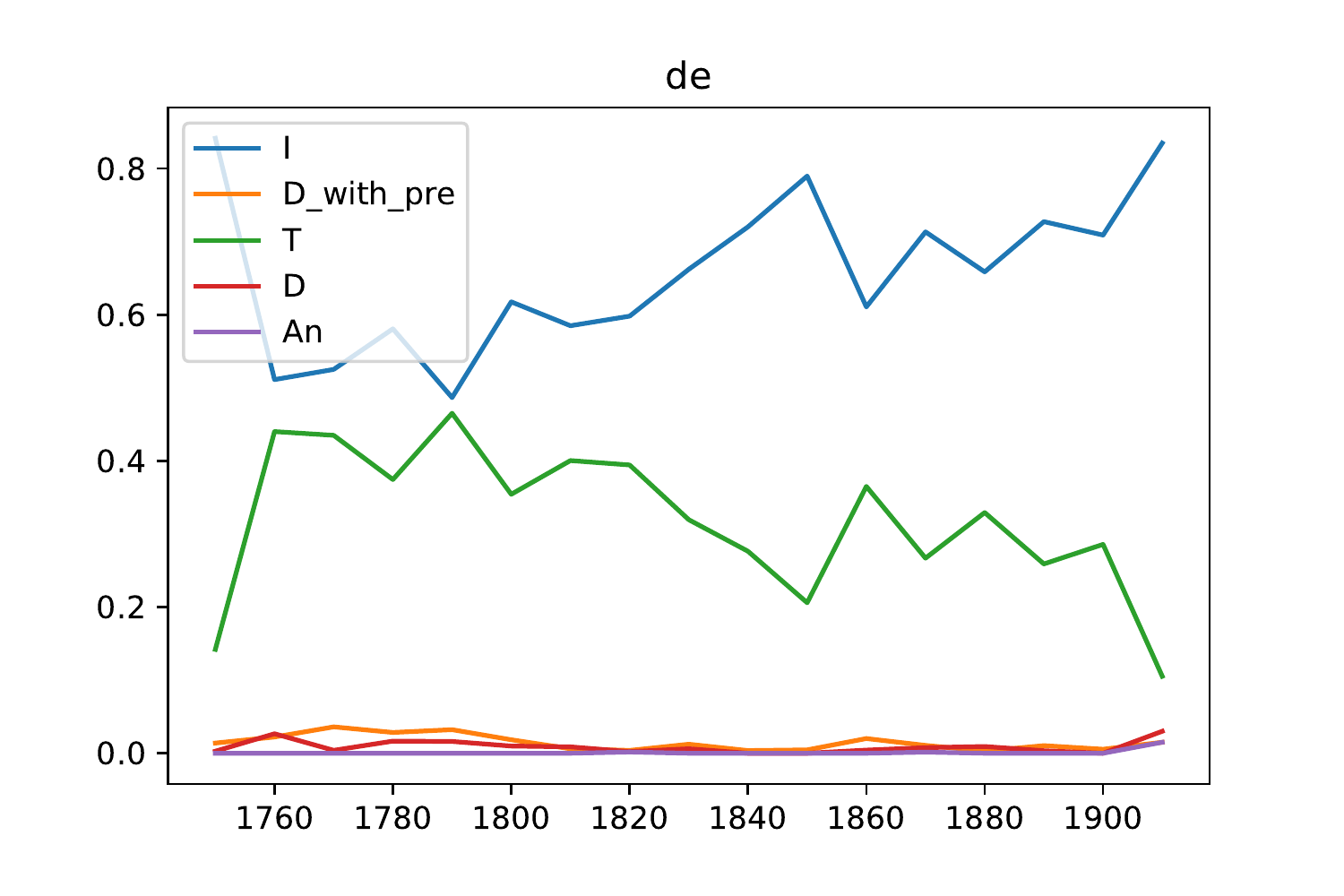}
        \vspace{-10mm}
    \end{subfigure}\\
    \begin{subfigure}[b]{0.45\textwidth}
        \centering
        \includegraphics[width=\textwidth]{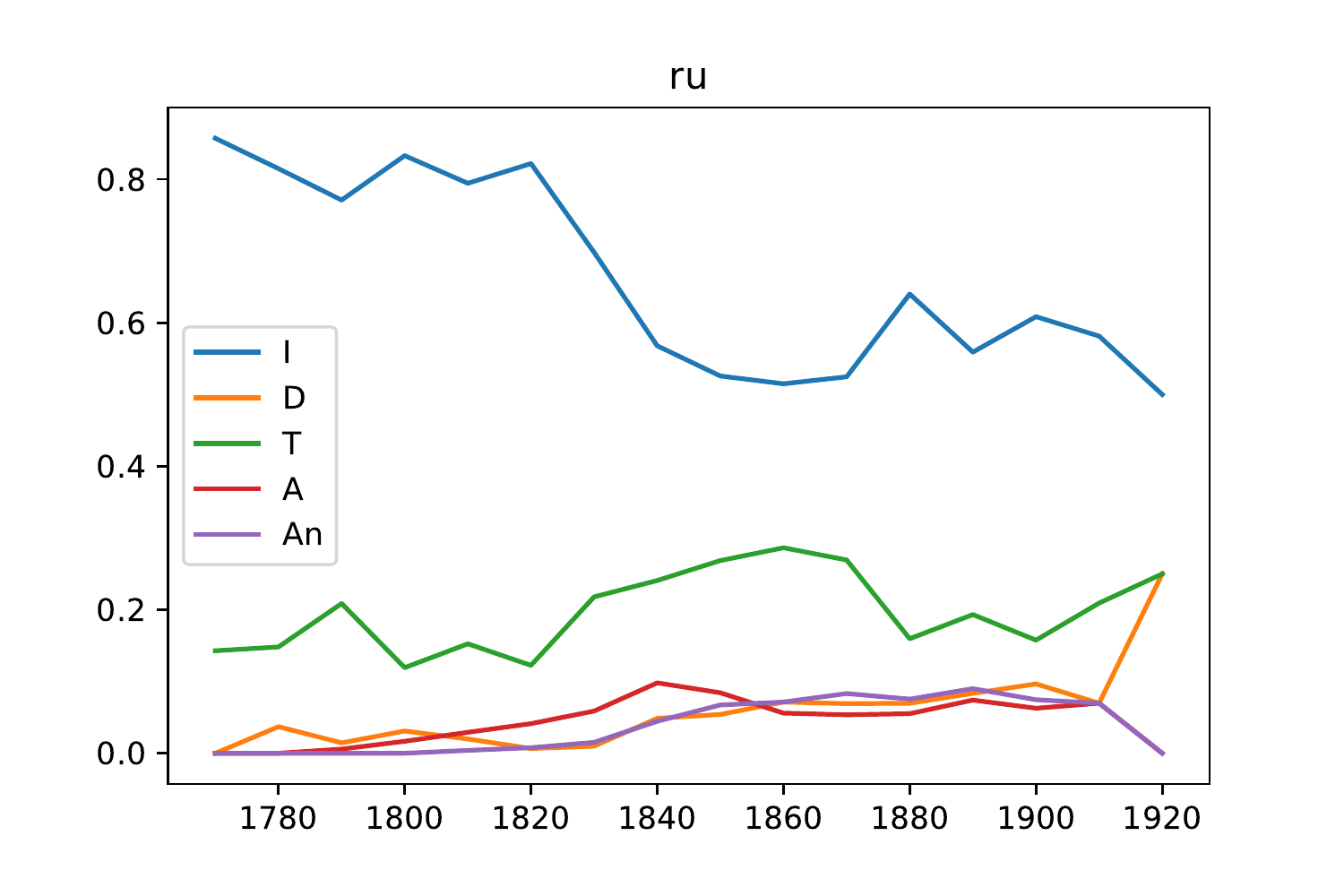}
        \vspace{-10mm}
    \end{subfigure}
    \hspace{1em}
    \begin{subfigure}[b]{0.45\textwidth}
        \centering
        \includegraphics[width=\textwidth]{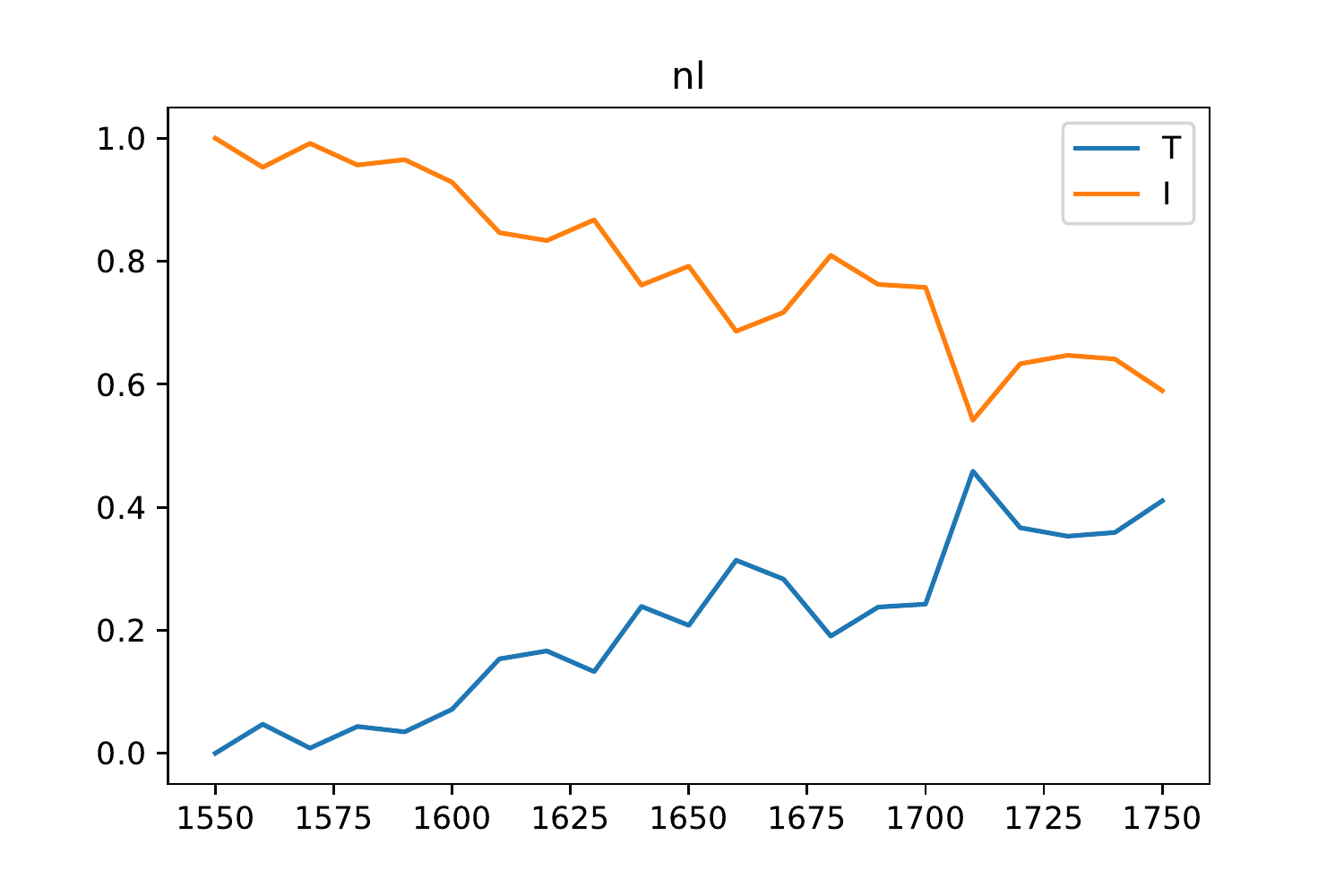}
        \vspace{-10mm}
    \end{subfigure}\\
    \begin{subfigure}[b]{0.45\textwidth}
        \centering
        \includegraphics[width=\textwidth]{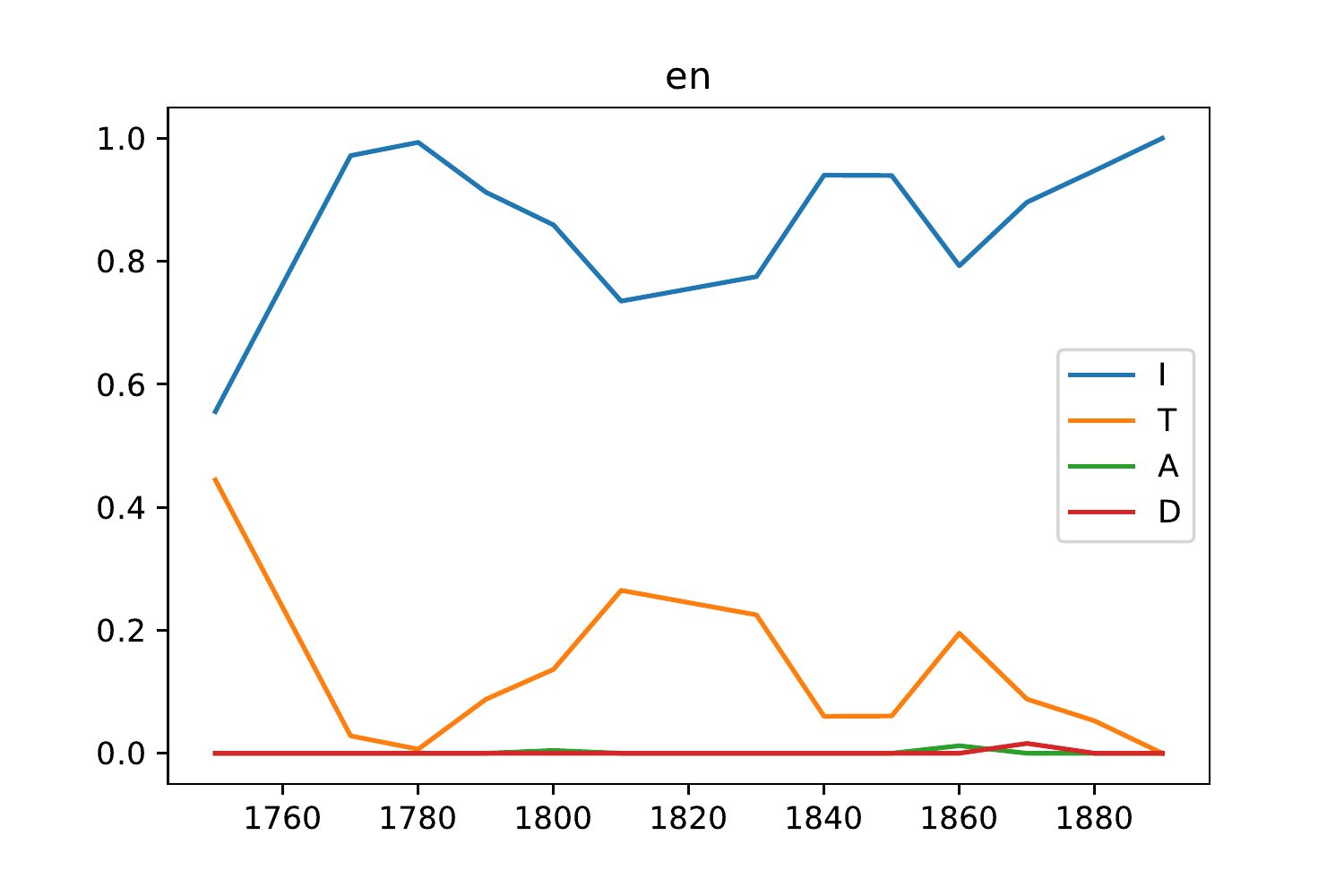}
    \end{subfigure} 
    \hspace{1em}
    \caption{Historical changes in the use of meters and proportion of given forms in the corpora. In the Czech works,we see a radical shift from trochee- to iamb-based works. There is also a gradual rise in trochee usage in the Dutch song corpus. This likely reflects the increasing representation of printed folklore texts}
    \label{S4_Fig}
\end{figure}

\clearpage
\subsection*{S2 Appendix}\label{S2_Appendix}
{\bf Metrical annotation} A poem's form may be recognized at different levels of granularity. Consider ``The Raven'' by Edgar Poe:  

\begin{enumerate}
    \item Generally, the metrical pattern of this poem is based on the trochee, i.e. it is made up of a recurring binary rhythmic unit (foot) in which a strong (stressed) syllable is followed by a weak (unstressed) one.
    
    \item Most lines in ``The Raven'' consist of eight trochaic feet: the overall meter is defined as trochaic octameter (shortened here to ``Trochee-8'').

    \item Readers of ``The Raven'' will notice that its octametric lines are organized into stable five-line units followed by a last (sixth) line that is shortened to tetrameter. At a stanzaic level, this form could be coded as Trochee-888884. Clearly it is so specific that any recurrence of it in other works signals a connection to Poe's poem. 
    
    \item This formal description could be expanded to include rhyme patterns. For the final rhymes in ``The Raven'', the coding would be: Trochee-888884-ABCBBB. Research suggests that in some cases (e.g. cultural borrowings from the syllabic versification to the accentual-syllabic traditions),  distinctive rhyme schemes may be associated with semantic traditions independently of meter. \cite{polilova_spanish_2018}.
    
    \item Finally, the rhythm of poems is sensitive to the choice of rhyme type, i.e. whether a rhyme ends on a stressed syllable (an acatalectic or masculine rhyme) or an unstressed one (a catalectic or feminine rhyme). When this factor is taken into account,the coding of ``The Raven'' would look like Trochee-888884-ABCBBB-fmfmmm.
\end{enumerate}

\begin{figure}
    \centering
    \begin{subfigure}[b]{0.05\textwidth}
        \centering
        \caption{}
        \label{fig:schemes_cs}
        \vspace{5cm}
    \end{subfigure}    
    \begin{subfigure}[b]{0.4\textwidth}
        \centering
        \includegraphics[width=\textwidth]{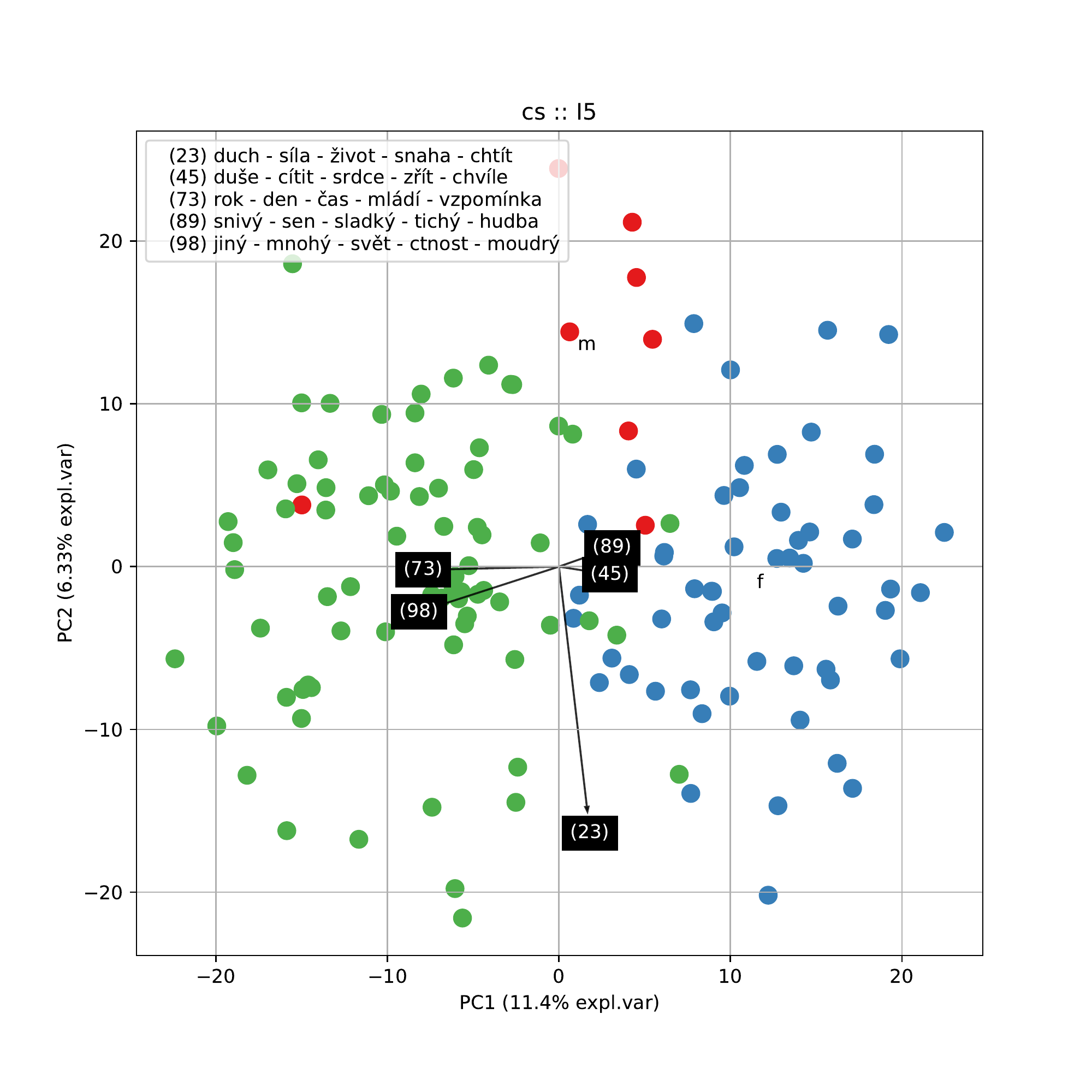}
        \vspace{-10mm}
    \end{subfigure}    
    \hspace{1em}
    \begin{subfigure}[b]{0.05\textwidth}
        \centering
        \caption{}
        \label{fig:schemes_de}
        \vspace{5cm}
    \end{subfigure}
    \begin{subfigure}[b]{0.4\textwidth}
        \centering
        \includegraphics[width=\textwidth]{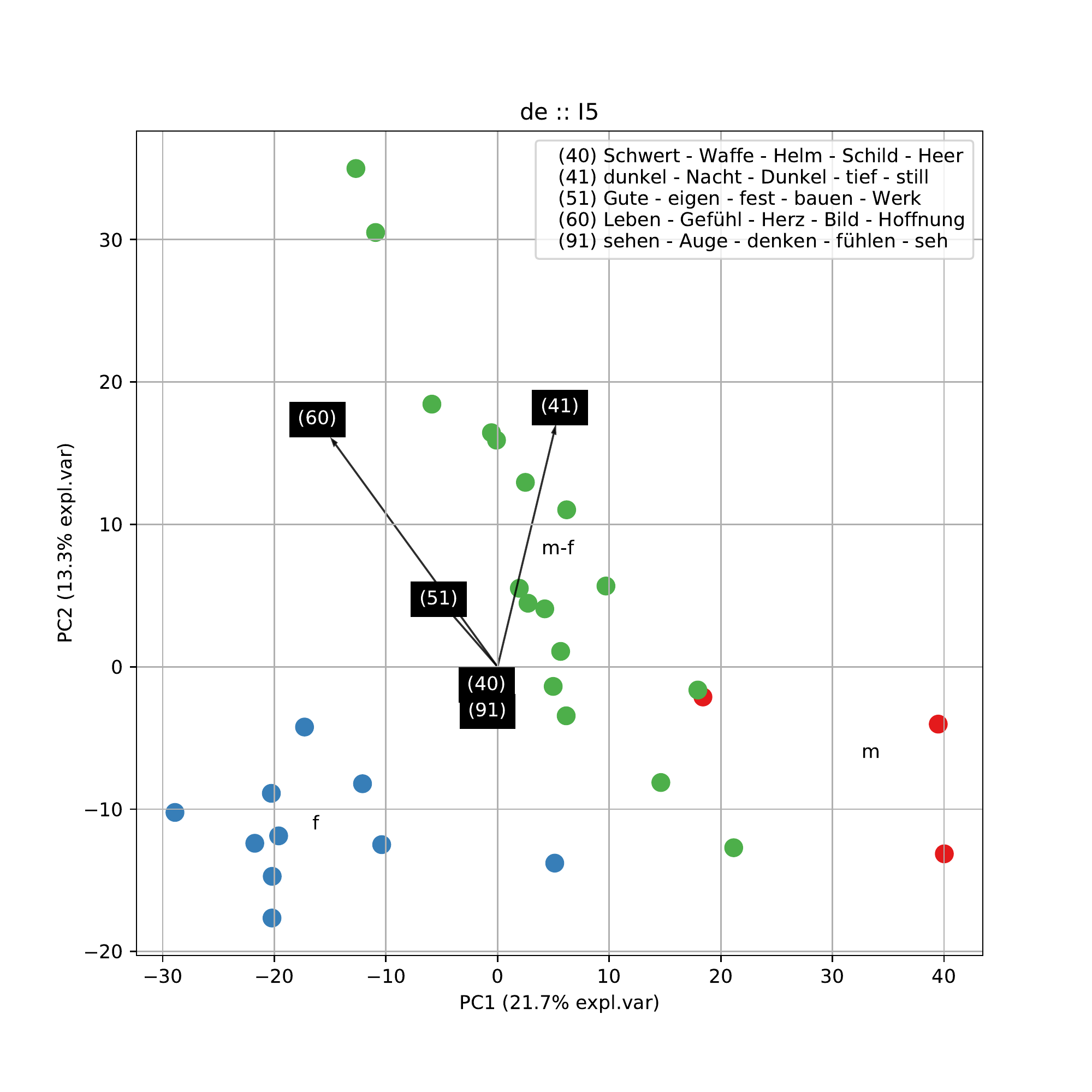}
        \vspace{-10mm}
    \end{subfigure}\\
    \caption{PCA biplots of Czech (\subref{fig:schemes_cs}) and German (\subref{fig:schemes_de}) variants of iambic pentameter with eigenvectors of the five most influential topics. Single random sampling. Each variant is defined as the shortest reoccurring pattern of line endings within the poem (i.e. whether these are masculine, feminine or dactylic)}
    \label{S5_Fig}
\end{figure}

As this summary shows, form may be seen as hierarchically organized: variations may occur from the level of the most abstract pattern (the trochee) through to the specific implementation of a meter in a highly regularized stanza. In this study, we aim for the mid-level of this hierarchy (number 2 on the list above). We apply a metrical type that is specific enough to register as structurally different but also abstract enough to be reasonably represented across the corpora and resistant to annotation errors and inconsistencies. This means that our analysis focuses on general relationships among meters but is not sensitive to possible variations within a particular form. Clearly the semantic representation of this ``general'' metrical type will be biased towards its most frequent metrical arrangements. Supplementary Fig. \ref{S5_Fig} shows that the most common metrical variants within a metrical type (e.g. different variants of Iamb-5) also remain semantically recognizable.


\begin{table}
\centering
\begin{tabular}{lcccccccccc}
\hline
& \multicolumn{2}{c}{Czech} & \multicolumn{2}{c}{German} & \multicolumn{2}{c}{Russian} & \multicolumn{2}{c}{Dutch} & \multicolumn{2}{c}{English} \\
\# of topics & median & IQR & median & IQR & median & IQR & median & IQR & median & IQR \\
\hline
20  & 0.57 & 0.05 & 0.72 & 0.12 & 0.39 & 0.05 & 0.92 & 0.23 & 1 & 0\\
50  & 0.64 & 0.19 & 0.74 & 0.07 & 0.45 & 0.12 & 0.74 & 0.03 & 1 & 0\\
100 & 0.62 & 0.14 & 0.7  & 0.08 & 0.48 & 0.15 & 0.74 & 0.03 & 1 & 0\\
150 & 0.64 & 0.2  & 0.71 & 0.08 & 0.4  & 0.11 & 0.71 & 0.06 & 1 & 0\\
\hline
\end{tabular}
\caption{Random 100-poem samples taken without replacement for each meter in vector spaces defined by LDA topic models(20, 50, 100, 150 topics). The results of our H1-related experiment show no qualitative variation regardless of the number of topics used to train an LDA model. We repeat the general ``semantic halo'' experiment in its full form (10,000 clustering iterations) for four different LDA models and report the Adjusted Rand Index mean along with the interquartile range.}
\label{S1_Table}
\end{table}


\begin{table}
\centering
\begin{tabular}{lcccccccccc}
\hline
\multicolumn{2}{c}{Czech} & \multicolumn{2}{c}{German} & \multicolumn{2}{c}{Russian} & \multicolumn{2}{c}{Dutch} & \multicolumn{2}{c}{English} \\
median & IQR & median & IQR & median & IQR & median & IQR & median & IQR \\
\hline
0.11 & 0.02 & 0.42 & 0.11 & 0.10 & 0.04 & 0.12 & 0.05 & 0.56 & 0.27\\
\hline
\end{tabular}
\caption{Random 100-poem samples taken without replacement for each meter in vector spaces defined by part-of-speech frequencies. Adjusted Rand Index of $k$-means clustering.  Accentual-syllabic meters are systems of limitations superimposed on language. As such, they transform natural morphological and syntactical affordances \cite{gasparov_linguistics_2008}. This is why we can distinguish poetry from prose so easily based on word frequencies \cite{storey_like_2020}. In addition, the distribution of parts of speech differs across metrical forms. Some words are more common in ternary meters simply because they are less likely to appear in binary meters for prosodic reasons. This has little connection with the semantics of meter but instead reflects the structural properties of verse. To make sure our pre-processing steps mitigated the problem of morphological differences, we repeat the clustering procedure from the set-up of the first experiment (H1). Here we use the frequencies of parts of speech that were included in LDA model (nouns, adjectives, verbs) as a feature set. Table shows that accuracy is significantly lower in this case than when clustering is topic-based.}
\label{S2_Table}
\end{table}


\begin{sidewaystable}
\centering
\begin{tabular}{cccc}

\hline
\multicolumn{4}{c}{Czech} \\
I4 & I5 & T4 & T5  \\
\hline
 kniha, psát, verš &
 duch, síla, život &
 blaho, blahý, srdce &
 vlast, národ, český\\

 ňadro, hruď, srdce &
 kniha, psát, verš &
 bůh, ctnost, svět &
 král, trůn, říše\\

 jaro, jarní, květ &
 velký, věk, lidstvo &
 bůh, dík, Toba &
 pravit, dít, děva\\

 vědět, povědět, dít &
 duše, cítit, srdce &
 dobrý, pan, říkat &
 pravda, slovo, řeč\\

 rok, den, čas &
 tvář, zrak, oko &
 mladý, hoch, dívka &
 jiný, mnohý, svět\\

\hline
\multicolumn{4}{c}{German} \\
I3 & I4 & I5 & T4  \\
\hline

 Wald, Berg, Feld &
 rufen, sehen, kommen &
 können, kommen, sehen &
 Baum, Zweig, grün\\

 Dan, welt, loben &
 laufen, tanzen, Kopf &
 leise, lauschen, stehen &
 Bruder, begraben, Muss\\

 Kind, wär, arm &
 Mann, Weib, Frau &
 Flamme, Feuer, Glut &
 schön, Venus, zieren\\

 Baum, Zweig, grün &
 sprechen, sagen, tun &
 Leben, Gefühl, Herz &
 Schönheit, hold, lieblich\\

 werd, ehren, würd &
 gut, Geld, schlecht &
 hoch, Flügel, fliegen &
 einen, Einer, rasch\\

\hline
\multicolumn{4}{c}{Russian} \\
I4 & I5 & I6 & T4  \\
\hline
 vladyka, bog, gospod' &
 hram, venets, altar' &
 dar, dostojnyj, serdtse &
 knjaz', rus', vitjaz'\\

 davat', mysl', slovo &
 den', drug, moch' &
 roscha, holm, les &
 chasha, vino, pir\\

 poet, pevets, muza &
 ljubov', strast', serdtse &
 uzhasnyj, strashnyj, smert' &
 deva, junyj, zhenih\\

 nadezhda, dusha, zhizn' &
 dusha, chuvstvo, serdtse &
 pravo, chin, primer &
 ruchej, krylo, ptitsa\\

 vdohnovenie, vostorg, mechta &
 vdohnovenie, vostorg, mechta &
 pravda, dobro, zlo &
 zoloto, stena, pyshnyj\\

\hline
\multicolumn{4}{c}{Dutch} \\
I3 & I4 & I5 & T4  \\
\hline
 lief, mogen, pijn &
 iesu, jesu, christum &
 dij, all, as &
 min, zin, hebben\\

 zeer, hebben, doen &
 vader, geven, geest &
 heilig, heiligen, geest &
 wijn, drinken, bier\\

 gods, woord, worden &
 bloed, vlees, ziel &
 heer, hebben, zullen &
 zult, zullen, geven\\

 moeten, tijd, mogen &
 sijt, beminde, getal &
 end, wt, zullen &
 zoet, kus, kussen\\

 satan, sint, godes &
 kind, moeder, klein &
 du, mit, dij &
 vrolijk, laten, hond\\
 
 \hline
\multicolumn{4}{c}{English} \\
I4 & I5  \\
\hline
 knight, lord, hall &
 sick, find, whole\\

 harp, strain, bard &
 new, good, old\\

 spear, horse, steed &
 great, tree, small\\

 roof, tower, pile &
 mind, kind, hail\\

 word, speak, answer &
 certain, blame, glide\\

\hline
\end{tabular}
\caption{Distinctive topics for the most common meters. The most distinctive topics for the most common meters. For each meter, topic probabilities are averaged across the entire corpus. These values are transformed into $z$-scores across particular meters. The table shows the five topics with the highest $z$-scores for each meter.}
\label{tab:ari_periods}
\end{sidewaystable}

\clearpage

\section*{Acknowledgements}

This study was supported by the Czech Science Foundation  as part of project no. 20-15650S. We would like to thank Klemens Bobenhausen, Benjamin Hammerich, Arthur Jacobs and Yurii Zelenkov for their help with data acquisition and annotation.

\printbibliography
\end{document}